%% file: main.tex
\documentclass{article}

\usepackage{arxiv}

\usepackage[utf8]{inputenc} 
\usepackage[T1]{fontenc}    
\usepackage{url}            
\usepackage{booktabs}       
\usepackage{amsfonts}       
\usepackage{nicefrac}       
\usepackage{microtype}      
\usepackage{lipsum}
\usepackage{amsmath}
\usepackage{amsthm}
\usepackage{amssymb}   
\usepackage{bigints}
\usepackage{bm}
\usepackage{diagbox}
\usepackage{makecell}
\usepackage{multirow}
\usepackage{caption}
\usepackage{subcaption}
\usepackage{graphicx}
\usepackage[ruled,vlined]{algorithm2e}
\include{pythonlisting}
\usepackage{xcolor}
\usepackage{mathtools}
\usepackage{enumitem}

\usepackage{hyperref}
\hypersetup{
    colorlinks=true,
    citecolor=black,
    linkcolor=black,
    filecolor=black,
    urlcolor=black,
}

\newcommand{\R}{\mathbb{R}}

\theoremstyle{definition}

\numberwithin{equation}{section}

\theoremstyle{plain}

\title{Long-time integration of parametric evolution equations with physics-informed DeepONets}

\author{
  Sifan Wang \\
  Graduate Group in Applied Mathematics \\
  and Computational Science \\
  University of Pennsylvania\\
  Philadelphia, PA 19104 \\
  \texttt{sifanw@sas.upenn.edu} \\
   \And
  Paris Perdikaris \\
  Department of Mechanichal Engineering \\
  and Applied Mechanics\\
  University of Pennsylvania\\
  Philadelphia, PA 19104 \\
  \texttt{pgp@seas.upenn.edu} \\
}

\begin{document}
\maketitle

\begin{abstract}
Ordinary and partial differential equations (ODEs/PDEs) play a paramount role in analyzing and simulating complex dynamic processes across all corners of science and engineering. In recent years machine learning tools are aspiring to introduce new effective ways of simulating PDEs, however existing approaches are not able to reliably return stable and accurate predictions across long temporal horizons. We aim to address this challenge by introducing an effective framework for learning  infinite-dimensional operators that map random initial conditions to associated PDE solutions within a short time interval. Such latent operators can be parametrized by deep neural networks that are trained in an entirely self-supervised manner without requiring any paired input-output observations. Global long-time predictions across a range of initial conditions can be then obtained by iteratively evaluating the trained model using each prediction as the initial condition for the next evaluation step. This introduces a new approach to temporal domain decomposition that is shown to be effective in performing accurate long-time simulations for a wide range of parametric ODE and PDE systems, from wave propagation, to reaction-diffusion dynamics and stiff chemical kinetics, all at a fraction of the computational cost needed by classical numerical solvers. 
\end{abstract}

\keywords{Deep learning \and Computational science \and Differential equations \and  Dynamical systems}

\section{Introduction}

Evolution equations, typically expressed as systems of time-dependent ordinary or partial differential equations (ODEs/PDEs), play a prominent role in the analysis, modeling and simulation of complex dynamical systems across diverse scientific domains, from fluid mechanics, to electromagnetics, quantum mechanics, and elasticity \cite{courant2008methods}. Classical approaches to simulating such equations often assume a discrete, finite-dimensional representation of the unknown solution (typically parametrized by a linear combination of fixed features such as polynomials, trigonometric functions, etc.), the weights of which are inferred by solving large linear or nonlinear systems, depending on the nature of the underlying equation and the scheme used to to discretize it  \cite{moin2010fundamentals}. For time-dependent problems these weights vary with time, and can be iteratively updated via appropriate time-integration schemes that discretize the temporal prediction horizon into a small number of steps, the size of which is dictated by the governing time-scales of the problem and the stability properties of the temporal discretization employed \cite{iserles2009first}. This general workflow has been thoroughly studied over the last several decades, leading to robust and provably accurate techniques such as the finite-element method \cite{hughes2012finite}, Runge-Kutta schemes \cite{iserles2009first}, and Krylov subspace methods \cite{saad2003iterative} that serve as the main workhorses of modern computational science and engineering. But as the complexity of the underlying evolution equations increases, so does the cost of simulating them; a cost that quickly becomes unbearable when multiple scenarios need to be queried (e.g., corresponding to different initial and boundary conditions (IBCs), random inputs, forcing terms, etc.), and/or when multi-scale interactions dictate the underlying dynamics.

As the machine learning (ML) revolution is persistently reaching all corners of science, a new wave of techniques are being proposed for accelerating the simulation of ODEs and PDEs \cite{karniadakis2021physics}. Instead of representing the target solution using a set of fixed features that are determined a-priori, neural networks \cite{psichogios1992hybrid,lagaris1998artificial} and kernel machines \cite{raissi2017inferring, chen2021solving} offer the possibility of learning effective representations that are adapted to the underlying evolution law. A representative example is the framework of physics-informed neural networks (PINNs) \cite{raissi2019deep} that opts to represent the entire spatio-temporal solution of a PDE system using a single deep neural network that is trained to jointly fit observed data (e.g., IBCs), as well as ensure that the predicted solution satisfies the underlying system of PDEs by minimizing its residual. An attractive property of this approach is that it no longer requires a spatial or temporal discretization of the PDE, nor it requires any external training data (other than knowledge of appropriate IBCs). Moreover, the entire global solution can be rapidly obtained once the network has been trained \cite{raissi2019deep}. However, this remarkable flexibility often comes at the price of reduced accuracy, as well as a multitude of caveats that hinder the training and convergence of such models \cite{wang2020understanding, wang2020and, wang2020eigenvector}. Nevertheless, PINNs \cite{raissi2019physics}, their variants \cite{kharazmi2019variational,jagtap2020extended,meng2020ppinn}, and other ML-based approaches \cite{zhu2019physics,geneva2020modeling,sanchez2020learning} are currently enjoying increased popularity across diverse applications including fluid mechanics \cite{raissi2020hidden,tartakovsky2020physics}, heat transfer \cite{hennigh2020nvidia,cai2021physics}, bio-engineering \cite{kissas2020machine,sahli2020physics}, materials \cite{lu2020extraction,chen2020physics,goswami2020transfer}, and finance \cite{elbrachter2018dnn,han2018solving}. However, a major challenge still remains unsolved and has been largely overlooked in the existing literature: ML-based approaches often fail to accurately simulate evolution equations over a long-time prediction horizon.

In this work we attempt to address this fundamental challenge by leveraging the recently developed framework of physics-informed deep operator networks (DeepOnets) \cite{wang2021learning} to parametrize and learn the solution operator that maps random initial conditions to their associated ODE/PDE solutions within a short time interval. The proposed deep learning model can be trained in an entirely self-supervised manner (i.e. without the need for any paired input-output data), only assuming knowledge of the ODE/PDE model form and its associated IBCs. Once the model has been trained on a collection of initial conditions, it can be used to construct the global ODE/PDE solution across a long-time prediction horizons via a simple iterative procedure in which the model prediction over a short time-step is used as an initial condition for the next evaluation. We demonstrate that this  approach can effectively enable the long-time integration of evolution equations subject to a range of initial conditions with good generalization accuracy, all at a fraction of the computational cost needed by classical numerical solvers. Taken together, the computational infrastructure developed in this work can have a broad technical impact in significantly reducing computational costs and accelerating scientific modeling of complex non-linear, non-equilibrium processes across diverse applications.

The remaining of this paper is structured as follows. In section \ref{sec: PINNs}, we provide an overview of the PINNs framework put forth by Raissi {\it et al.} \cite{raissi2019deep} and demonstrate its fundamental limitations in approximating ODE/PDE solutions over long-time horizons. Section \ref{sec: Methods} provides a detailed discussion of our main technical contributions, starting with a recap on physics-informed DeepOnets \cite{wang2021learning} in section \ref{sec: DeepOnet}, followed by the proposed formulation for tackling long-time integration problems in section \ref{sec: integration}. Further, in section \ref{sec: results} we present a series of comprehensive numerical studies to assess the performance of the proposed long-time integration framework across a range of parametric ODE/PDE systems involving wave propagation, reaction-diffusion dynamics, and stiff chemical kinetics. Finally, section \ref{sec:discussion} concludes with a discussion of our main findings, potential pitfalls, and shortcomings, as well as future research directions emanating from this study. All code and data accompanying this manuscript will be made available at \url{https://github.com/PredictiveIntelligenceLab/Long-time-Integration-PI-DeepONets}.

\section{Physics-informed neural networks}

\label{sec: PINNs}

In this section, we give a brief review of physics-informed neural networks (PINNs) \cite{raissi2019physics} for solving time-dependent ODEs and PDEs. Generally, we consider initial–boundary value problems taking the form
\begin{align}
\label{eq: PDE}
     &\bm{s}_t + \mathcal{N}_{\bm{x}}[\bm{s}] = \bm{0}, \quad  \bm{x} \in \Omega, t \in [0, T] \\
     \label{eq: BC}
    &\bm{s}(\bm{x}, t) = \bm{g}(\bm{x}, t), \quad  \bm{x} \in \partial \Omega, t \in [0, T] \\
         \label{eq: IC}
     &\bm{s}(\bm{x}, 0) = \bm{u}(\bm{x}), \quad \bm{x} \in \Omega,
\end{align}
where $\bm{x}$ and $t$ represent spatial and temporal coordinates, respectively,  $\mathcal{N}_{\bm{x}}$ denotes a differential operator with respect to $\bm{x}$, and $\Omega \subset \R^n$ is an open, bounded domain with a well-behaved boundary $\partial \Omega$. In addition, $\bm{s}: \overline{\Omega} \rightarrow \R^m$ denotes the unknown latent quantity of interest that is governed by the PDE system of equation (\ref{eq: PDE}).

We proceed by approximating $\bm{s}(\bm{x})$ by a deep neural network $\bm{s}_{\bm{\theta}}(\bm{x})$, where $\bm{\theta}$ denotes all trainable parameters of the networks. Then, we can define the corresponding PDE residual as 
\begin{align}
\label{eq: residual}
    \bm{r}_{\bm{\theta}}(\bm{x}, t) := \frac{\partial}{\partial t}\bm{s}_{\bm{\theta}}(\bm{x},t) + \mathcal{N}_{\bm{x}}[\bm{s}_{\bm{\theta}}(\bm{x},t)],
\end{align}
where the partial derivatives of the neural network representation with respect to space and time coordinates can be readily computed to machine precision using forward or reverse mode automatic differentiation \cite{baydin2018automatic}. A physics-informed neural network can be trained by minimizing the following composite loss function
\begin{align}
    \label{eq: PINN_loss}
    \mathcal{L}(\bm{\theta}) = \lambda_r \mathcal{L}_r(\bm{\theta}) + \lambda_{\text{bc}} \mathcal{L}_{\text{bc}}(\bm{\theta}) + \lambda_{\text{ic}} \mathcal{L}_{\text{ic}}(\bm{\theta}),
\end{align}
where 
\begin{align}
    \label{eq: loss_r}
    &\mathcal{L}_r(\bm{\theta}) = \frac{1}{N_r} \sum_{i=1}^{N_r} \left| \bm{r}_{\bm{\theta}}(\bm{x}_r^i, t_r^i) \right|^2, \\
    \label{eq: loss_bc}
     &\mathcal{L}_{\text{bc}}(\bm{\theta}) = \frac{1}{N_{\text{bc}}} \sum_{i=1}^{N_{\text{bc}}} \left| \bm{s}_{\bm{\theta}}(\bm{x}_{\text{bc}}^i, t_{\text{bc}}^i) - \bm{g}(\bm{x}_{\text{bc}}^i, t_{\text{bc}}^i) \right|^2, \\
    \label{eq: loss_ic}
     &\mathcal{L}_{\text{ic}}(\bm{\theta}) = \frac{1}{N_{\text{ic}}} \sum_{i=1}^{N_{\text{ic}}} \left| \bm{s}_{\bm{\theta}}(\bm{x}_{\text{ic}}^i, 0) - \bm{u}(\bm{x}_{\text{ic}}^i) \right|^2.
\end{align}
Here,  $N_r$, $N_{\text{bc}}$ and $N_{\text{ic}}$ denote the batch-sizes of the "training data" $\{(\bm{x}_{r}^i, t_{r}^i),  \bm{f}(\bm{x}_{r}^i, t_{r}^i)\}_{i=1}^{N_{r}}$, $\{(\bm{x}_{\text{bc}}^i, t_{\text{bc}}^i),  \bm{g}(\bm{x}_{\text{bc}}^i, t_{\text{bc}}^i)\}_{i=1}^{N_{\text{bc}}}$ and $\{\bm{x}_{\text{ic}}^i, \bm{h}(\bm{x}_{\text{ic}}^i)\}_{i=1}^{N_{\text{ic}}}$, respectively, which are randomly sampled in the computational domain and the boundary at each iteration of a stochastic gradient descent algorithm. 
Moreover, the parameters $\left\{ \lambda_r,  \lambda_{\text{bc}}, \lambda_{\text{ic}} \right\}$ correspond to weight coefficients in the loss function that can effectively assign a different learning rate to each individual loss term. These weights may be user-specified or tuned automatically during network training \cite{wang2020understanding, wang2020and, mcclenny2020self}. 

\begin{figure}
    \centering
    \includegraphics[width=0.6\textwidth]{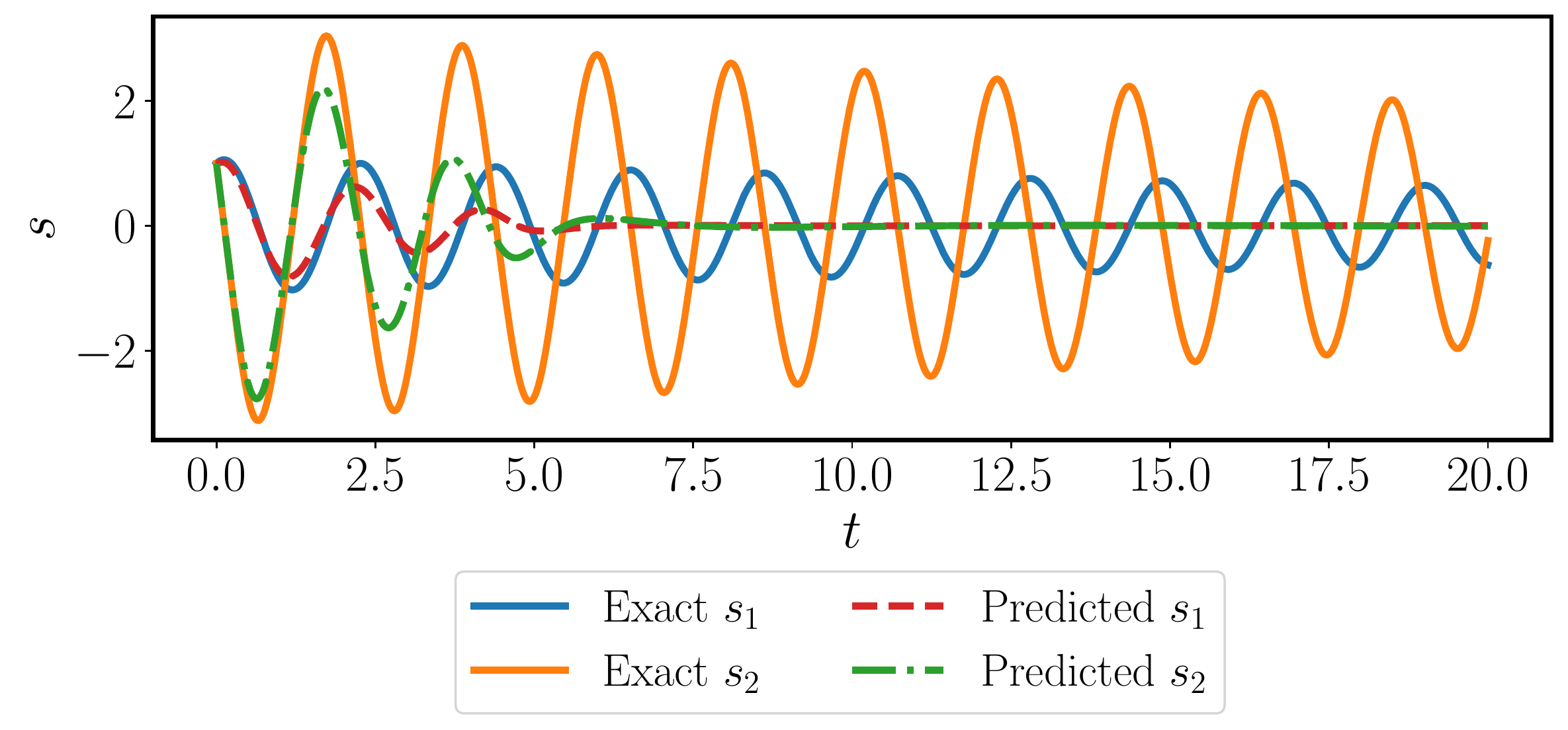}
    \caption{{\em Gravity Pendulum:} Predicted solution for $s_1(t)$ and $s_2(t)$  versus the corresponding reference solution. The result is obtained by training a conventional PINN (5 layers, 100 hidden units, tanh activations) for $10^5$ iterations of gradient descent using the Adam optimizer. Evidently, the model predictions collapse after $t=10$.}
    \label{fig: PINN_pendulum_s_pred}
\end{figure}

Despite a series of promising results \cite{raissi2020hidden,kissas2020machine, sahli2020physics, cai2021physics}, the original formulation of Raissi {\em et al.} \cite{raissi2019physics} typically fails to handle long-time prediction tasks. To illustrate this, let us consider a simple gravity pendulum with damping governed by the following ODE system
\begin{align}
    \label{eq: pendulum_1}
    &\frac{d s_1}{dt} =  s_2, \\
        \label{eq: pendulum_2}
    &  \frac{d s_2}{dt} = - \frac{b}{m} s_2  - \frac{g}{L} \sin(s_1),
\end{align}
for $ t \in [0, T]$. The initial condition is given by $s_1(0) = s_2(0) = 1$. In this example, we take $m = L = 1$, $b = 0.05$ and $g = 9.81$. We are interested in using PINNs to solve this two-dimensional ODE system up to $T = 20$. To this end, we approximate the latent variables $s_1, s_2$ by a 5-layer fully-connected neural network $\bm{s}_{\bm{\theta}} = [s^{(1)}_{\bm{\theta}}, s^{(2)}_{\bm{\theta}}]$
with 100 units per hidden layer, and define the ODE residual as
\begin{align}
    \label{eq: PINN_pendulum_res_1}
    &r^{(1)}_{\bm{\theta}}(t) = \frac{d s^{(1)}_{\bm{\theta}}(t)}{ dt} - s^{(2)}_{\bm{\theta}}(t), \\
    \label{eq: PINN_pendulum_res_2}
    & r^{(2)}_{\bm{\theta}}(t) = \frac{d s^{(2)}_{\bm{\theta}}(t)}{ dt} + \frac{b}{m} s^{2}_{\bm{\theta}}(t) - \frac{g}{L} \sin(s^{(1)}_{\bm{\theta}}(t)).
\end{align}
The corresponding PINNs loss function is given by
\begin{align}
    \label{eq: PINN_Pendulum_loss}
    \mathcal{L}(\bm{\theta}) = \mathcal{L}_r(\bm{\theta}) + \mathcal{L}_{\text{ic}}(\bm{\theta}), 
\end{align}
where
\begin{align}
    &\mathcal{L}_r(\bm{\theta}) = \frac{1}{N_r} \sum_{i=1}^{N_r} \left[ \left| \bm{r}_{\bm{\theta}}^{(1)}(t_r^i)  \right|^2 +  \left| \bm{r}_{\bm{\theta}}^{(2)}(t_r^i)  \right|^2  \right], \\
    &\mathcal{L}_{\text{ic}}(\bm{\theta})  =  \left|s^{(1)}_{\bm{\theta}}(0) - s_1(0)  \right|^2 + \left|s^{(2)}_{\bm{\theta}}(0) - s_2(0)  \right|^2.
\end{align}
We set $N_r = 10^4$, and all collocation points $\{t_r^i\}_{i=1}^{N_r}$ are randomly sampled in $[0, T]$ at each iteration during training. We train the network by minimizing the loss function (\ref{eq: PINN_Pendulum_loss}) for $10^5$ iterations of gradient descent using the Adam optimizer with default settings \cite{kingma2014adam}. 
A comparison of the predicted solutions against their corresponding numerical estimation obtained with a conventional adaptive Runge-Kutta solver \cite{iserles2009first} is shown in Figure \ref{fig: PINN_pendulum_s_pred}. It is clear that the PINN model predictions collapse to zero after $T = 10$, which suggests that PINNs may be incapable of yielding accurate solutions for long-time integration problems.

There are some reasons that may explain the poor predictions and model collapse. One could be saturated activation functions due to 
large values of the input coordinates. Another possible reason may be the inability of neural networks to approximate high-frequency and complex functions because of spectral bias \cite{rahaman2019spectral}. Admittedly, recent work has provided some remedies that can be directly applied to this case to improve model performance \cite{tancik2020fourier,wang2020eigenvector,wight2020solving,meng2020ppinn}. For example, to avoid saturation of activations, one can normalize the inputs such that they lie in a reasonable range, although this trick will lead to very small coefficients in the ODE system, and, consequently, to a singular perturbation problem \cite{weinan2011principles}. Other approaches include, but are not limited to, using Fourier feature embeddings \cite{tancik2020fourier,wang2020eigenvector}, as well as employing and training multiple individual networks in different temporal sub-domains \cite{meng2020ppinn, wight2020solving,du2021evolutional}.
The former typically requires some prior knowledge of the frequency content of the latent solution in order to properly initialize the model \cite{wang2020eigenvector}, while the latter approaches inevitably lead to a large computational cost. In the following sections, we present a simple yet effective strategy to solve long-time integration problems via physics-informed DeepONets \cite{wang2021learning}.

\section{Methods}\label{sec: Methods}

\subsection{A primer on physics-informed DeepONets}
\label{sec: DeepOnet}

Recently, Lu {\em et. al.} \cite{lu2021learning} proposed deep operator networks (DeepONets), which aim to learn abstract nonlinear operators mapping functions between infinite-dimensional Banach spaces. In follow up work, Wang {\em et. al.} \cite{wang2021learning} developed  physics-informed DeepONets, introducing an effective regularization mechanism for biasing the outputs of DeepONet models towards ensuring physical consistency. Here we present a brief overview of  physics-informed DeepONets with a special focus on time-dependent PDEs (see equations (\ref{eq: PDE}) - (\ref{eq: IC})). 

\begin{figure}
    \centering
    \includegraphics[width=0.8\textwidth]{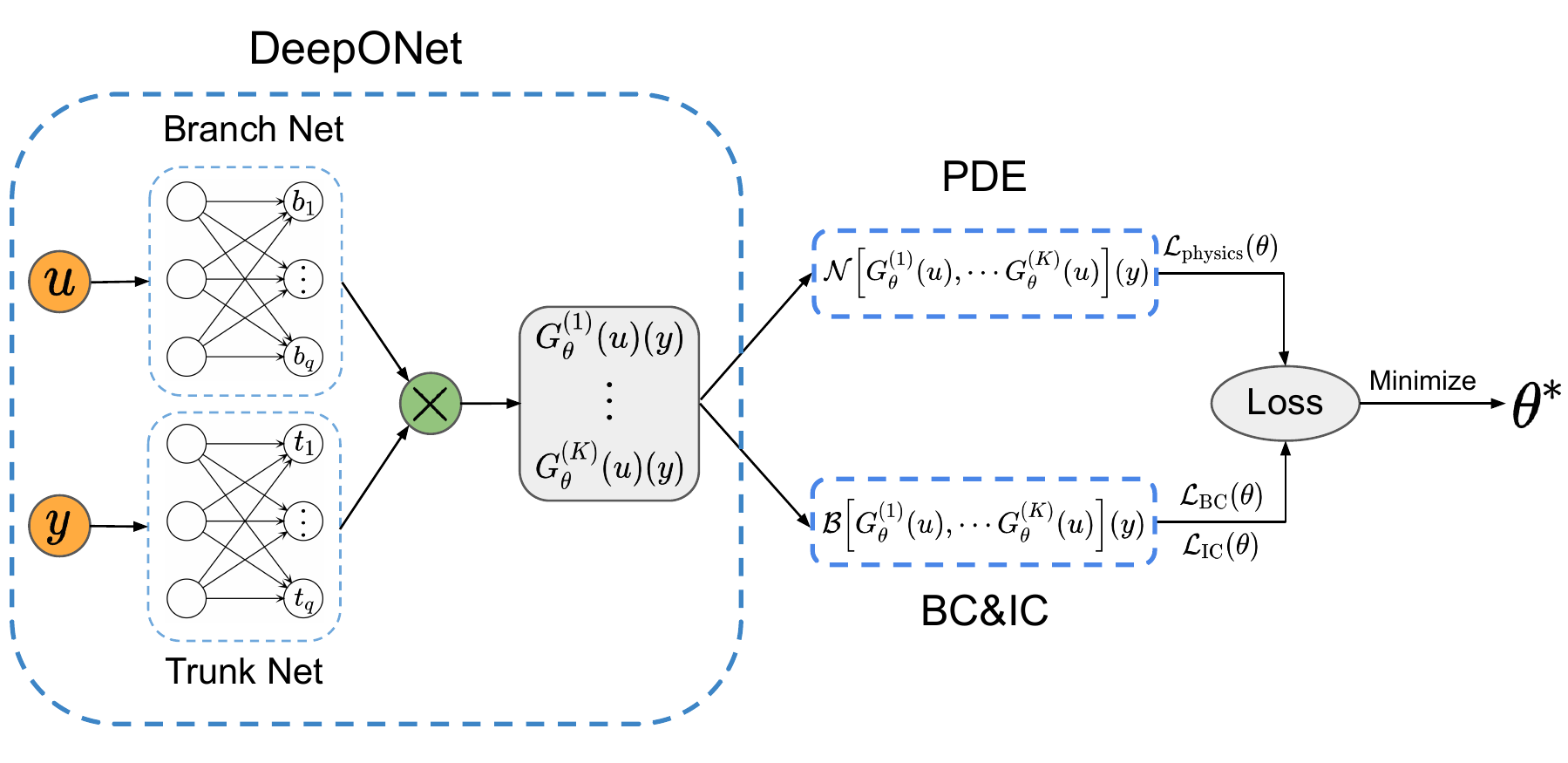}
    \caption{{\em Making DeepOnets physics-informed:}  The DeepONet architecture \cite{lu2021learning} consists of two sub-networks, the branch net for extracting latent  representations of input functions, and the trunk net for extracting latent representations of input coordinates at which the output functions are evaluated. A continuous and differentiable representation of the output functions is then obtained by merging the latent representations extracted by each sub-network via a dot product. Automatic differentiation can then be employed to formulate appropriate regularization mechanisms for biasing the DeepOnet outputs to satisfy a given system of PDEs.}
    \label{fig: deepOnet_architecture}
\end{figure}

Let $\mathcal{U}, \mathcal{S}$ be two separate Banach spaces.  We are interested in learning the solution operator $G$ from an initial condition $\bm{u} \in \mathcal{U}$ to the associated PDE solution $\bm{s}(\bm{x}, t) \in \mathcal{S}$.
To this end, we represent the solution map $G$ by a DeepONet $G_{\bm{\theta}}$. As illustrated in Figure \ref{fig: deepOnet_architecture},  the DeepONet architecture consists of two separate neural networks referred to as the "branch net" and "trunk net", respectively. The branch net  takes the parameters $\bm{s}$ as input and returns a features embedding $[b_1, b_2,\dots, b_q]^T \in \R^q$ as output, where $\bm{u} = [\bm{u}(\bm{x}_1), \bm{u}(\bm{x}_2), \dots, \bm{u}(\bm{x}_m) ]$ represents a function $\bm{u} \in \mathcal{U}$ evaluated at a collection of fixed locations $\{\bm{x}_i\}_{i=1}^m \subset \Omega$.  The trunk net takes the continuous coordinates $(\bm{x}, t)$ as inputs, and outputs a features embedding $[t_1, t_2,\dots, t_q]^T \in \R^q$. The final output of the DeepONet is obtained by merging the outputs of the branch and trunk networks via a dot product.  More specifically, a DeepONet $G_{\bm \theta}$ prediction of an input function $\bm{u}$ evaluated at $(\bm{x}, t)$  can be expressed by
\begin{align}
    \label{eq: deeponet_output}
    G_{\bm{\theta}}(\bm{u})(\bm{x}, t) = \sum_{k=1}^{q} \underbrace{b_{k}\left(\bm{u}\left(\bm{x}_{1}\right), \bm{u}\left(\bm{x}_{2}\right), \ldots, \bm{u}\left(\bm{x}_{m}\right)\right)}_{\text {branch }} \underbrace{t_{k}(\bm{x}, t)}_{\text {trunk }},
\end{align}
where $\bm{\theta}$ denotes the collection of all trainable weights and biases in the branch and trunk networks. Note that the outputs of a DeepONet model are continuously differentiable with respect to the query points $(\bm{x}, t)$, thus allowing us to employ automatic differentiation \cite{griewank1989automatic, baydin2018automatic} to compute the associated PDE residual 
\begin{align}
        \mathcal{R}_{\bm{\theta}}[\bm{u}](\bm{x}, t) =\frac{\partial G_{\bm{\theta}}(\bm{u})(\bm{x},t)}{ \partial t} +  \mathcal{N}_{\bm{x}}[G_{\bm{\theta}}(\bm{u})](\bm{x}, t).
\end{align}
Then, we can construct a physics-informed DeepONet by formulating the following loss function
\begin{align}
        \mathcal{L}(\bm{\theta}) = \mathcal{L}_{\text{ic}}(\bm{\theta}) + \mathcal{L}_{\text{bc}}(\bm{\theta}) +  \mathcal{L}_{r}(\bm{\theta}),
\end{align}
where
\begin{align}
      \mathcal{L}_{\text{ic}}(\bm{\theta}) &= \frac{1}{NP}\sum_{i=1}^N \sum_{j=1}^{P} \left|G_{\bm{\theta}}(\bm{u}^{(i)})(\bm{x}^{(i)}_{\text{ic}, j}, 0 ) - \bm{u}^{(i)}(\bm{x}^{(i)}_{\text{ic}, j})  \right|^2 \\
      \mathcal{L}_{\text{bc}}(\bm{\theta}) &= \frac{1}{NP}\sum_{i=1}^N \sum_{j=1}^{P} 
     \left|G_{\bm{\theta}}(\bm{u}^{(i)})(\bm{x}_{\text{bc}, j}^{(i)}, t^{(i)}_{\text{bc},j} ) - \bm{g}(\bm{x}^{(i)}_{\text{bc}, j}, t^{(i)}_{\text{bc},j} )\right|^2,\\
      \mathcal{L}_{r}(\bm{\theta}) &= \frac{1}{NQ}\sum_{i=1}^N \sum_{j=1}^{Q} \left|\mathcal{R}_{\bm{\theta}}[\bm{u}^{(i)}] (\bm{x}^{(i)}_{r,j}, t^{(i)}_{r, j})   \right|^2.
\end{align} 
Here $\{ \bm{u}^{(i)} \}_{i=1}^N$ denotes $N$ separate input functions sampled from $\mathcal{U}$. For each  $\bm{u}^{(i)}$, 
$\{(\bm{x}^{(i)}_{\text{ic},j}, 0\}_{j=1}^P$,  $\{(\bm{x}^{(i)}_{\text{bc},j}, t^{(i)}_{\text{bc},j})\}_{j=1}^P$ are $P$ locations sampled from $\Omega \times \{T = 0\}$ and $\partial \Omega \times [0, T]$ for enforcing the inital and boundary conditions, respectively. Besides,  $\{(\bm{x}^{(i)}_{\text{r},j}, t^{(i)}_{\text{r},j})\}_{j=1}^Q$ is a set of collocation points sampled from the computational domain $\Omega \times [0, T]$ for penalizing the parametric PDE residual. In contrast to the fixed sensor locations of $\{\bm{x}_i\}_{i=1}^m$, we remark that the locations of $\{(\bm{x}^{(i)}_{\text{ic},j}, 0\}_{j=1}^P$,  $\{(\bm{x}^{(i)}_{\text{bc},j}, t^{(i)}_{\text{bc},j})\}_{j=1}^P$ and  $\{(\bm{x}^{(i)}_{\text{r},j}, t^{(i)}_{\text{r},j})\}_{j=1}^Q$ may vary across different input samples $\bm{u}^{(i)}$.

\subsection{Long-time integration of evolution equations}
\label{sec: integration}

In this section, we present our main contribution for solving time-dependent PDEs (\ref{eq: PDE}) - (\ref{eq: IC}) involving long-time integration using physics-informed DeepONets. Instead of decomposing the temporal domain $[0, T]$ into many sub-domains and sequentially solving each short-time problem with an independent neural network, we train a single physics-informed DeepONet to learn the solution operator of the same PDE for a short time $t \in [0, \Delta t]$ subject to a distribution of initial conditions. Then we can obtain the inferred solution by recurrently replacing the initial condition with the model's prediction at $t  = \Delta t$, and evaluating again the forward pass of the trained model. The details of the proposed strategy are summarized in Algorithm  \ref{alg: long_time_integration}.

\begin{algorithm}
\SetAlgoLined
Suppose that $G_{\bm{\theta}}$ is a trained physics-informed DeepONet that learned the solution operator of the time-dependent time PDE (\ref{eq: PDE}) - (\ref{eq: IC}) for $t \in [0, \Delta t]$. Let $\{(\bm{x}_i, t_i)\}_{i=1}^P$ be a set of uniform grid points in $\Omega \times [0, \Delta t]$ and initialize the input function by $\bm{u}^0(\bm{x}) = \bm{u}(x)$ (the initial condition in equation (\ref{eq: IC})). \\
 \For{$k = 1, \dots, N$}{
    (a) Infer the solution by running the forward pass of the physics-informed DeepONet 
        \begin{align}
            \bm{s}^{k}(\bm{x}_i, t_i) = G_{\bm{\theta}}(\bm{u}^{k-1})(\bm{x}_i, t_i), \quad \text{for } i = 1, 2, \dots, P.
        \end{align}
   
  (b) Update the input function by
  \begin{align}
      \bm{u}^{k}(\bm{x}) = \bm{s}^{k}(\bm{x}, \Delta t)
  \end{align}

}
The final predicted solution in the whole domain $\Omega \times [0, T]$ can be obtained by concatenating all the inferred solutions $\{\bm{s}^{k}\}_{k=1}^{N}$.

\caption{Long-time integration of evolution equations with physics-informed DeepOnets.}
\label{alg: long_time_integration}
\end{algorithm}

\begin{figure}
    \centering
    \includegraphics[width=0.9\textwidth]{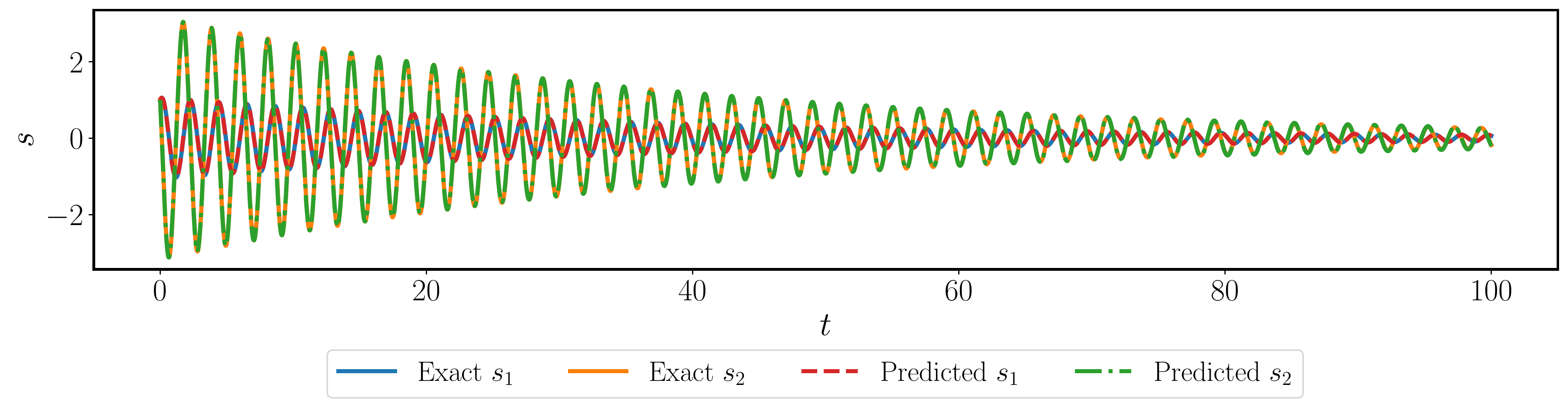}
     \caption{{\em Gravity Pendulum:} Predicted solution for $s_1(t)$ and $s_2(t)$  versus the corresponding reference solution. The result is obtained by applying Algorithm \ref{alg: long_time_integration} to a trained a physics-informed DeepONet.  
        The relative $L^2$ errors of $s_1$ and $s_2$ are $1.72\%$ and $1.63\%$, respectively. }
    \label{fig: PI_deeponet_pendulum_s_pred}
\end{figure}

To introduce more technical details, let us revisit the example of gravity pendulum presented in section \ref{sec: PINNs} and pursue its simulation up to $T = 100$. Before doing so, notice that the output of a vanilla DeepONet is a scalar, while the solution of the ODE system in equations (\ref{eq: pendulum_1}) - (\ref{eq: pendulum_2}) is a 2-dimensional vector.  To resolve this issue, we modify the forward pass (\ref{eq: deeponet_output}) such that the DeepONet output can be a vector. Specifically, suppose that a DeepONet outputs $n$ different scalar functions. Then the forward pass of $i$-th function is given by
\begin{align}
\label{eq: deeponet_multiple_outputs}
s_{\bm{\theta}}^{(i)} = 
G_{\bm{\theta}}^{(i)}(\bm{u})(\bm{x}, t) = \sum_{k=q_{i-1}+1}^{q_{i}} \underbrace{b_{k}\left(\bm{u}\left(\bm{x}_{1}\right), \bm{u}\left(\bm{x}_{2}\right), \ldots, \bm{u}\left(\bm{x}_{m}\right)\right)}_{\text {branch }} \underbrace{t_{k}(\bm{x}, t)}_{\text {trunk }}
\end{align}
for $i = 1, \dots, n$ where $0 = q_0 < q_1 < \cdots < q_{n} = q$.
For this 2D ODE, we take $n = 2$, $q = 200$ and $q_1 = 100$. Now we employ a DeepONet $G_{\bm{\theta}} = [G_{\bm{\theta}}^{(1)}, G_{\bm{\theta}}^{(2)}]$ to represent the solution map from initial conditions to the associated solutions in $[0, 1]$, where both the branch and trunk networks are 8-layer fully-connected neural networks with 100 units per hidden layer. Similar to equation (\ref{eq: PINN_pendulum_res_1}) - (\ref{eq: PINN_pendulum_res_2}), we can define the ODE residual for the physics-informed DeepONet model as
\begin{align}
    \label{eq: PI_deeponet_pendulum_res_1}
    &\mathcal{R}_{\bm{\theta}}^{(1)}[\bm{u}](t) = \frac{d G^{(1)}_{\bm{\theta}}(\bm{u})(t)}{ dt} - G^{(2)}_{\bm{\theta}}(\bm{u})(t), \\
    \label{eq: PI_deeponet_pendulum_res_2}
    & \mathcal{R}_{\bm{\theta}}^{(2)}[\bm{u}](t) =  \frac{d G^{(2)}_{\bm{\theta}}(\bm{u})(t)}{ dt} + \frac{b}{m} G^{2}_{\bm{\theta}}(\bm{u})(t) - \frac{g}{L} \sin(G^{(1)}_{\bm{\theta}}(\bm{u})(t)).
\end{align}
Then, the trainable parameters $\bm{\theta}$ can be optimized by minimizing the following loss
\begin{align}
        \mathcal{L}(\bm{\theta}) = \mathcal{L}_{\text{ic}}(\bm{\theta}) + \mathcal{L}_{r}(\bm{\theta}),
\end{align}
where
\begin{align}
      \mathcal{L}_{\text{ic}}(\bm{\theta}) &= \frac{1}{N}\sum_{i=1}^N  \left[ \left|G_{\bm{\theta}}^{(1)}(\bm{u}^{(i)})( 0 ) - u^{(i)}_1 \right|^2 +  \left|G_{\bm{\theta}}^{(2)}(\bm{u}^{(i)})( 0 ) - u^{(i)}_2 \right|^2 \right] \\
      \mathcal{L}_{r}(\bm{\theta}) &= \frac{1}{NQ}\sum_{i=1}^N  \sum_{j=1}^{Q} \left[ \left|\mathcal{R}^{(1)}_{\bm{\theta}}[\bm{u}^{(i)}] (t^{(i)}_{j})   \right|^2 + \left|\mathcal{R}^{(2)}_{\bm{\theta}}[\bm{u}^{(i)}] (t^{(i)}_{j})   \right|^2 \right].
\end{align} 
Here, $\bm{u}^{(i)} = [u_1^{(i)}, u_2^{(i)}]$ are the inputs of the branch network which denote the initial conditions. Moreover, for each $i$, $\{t^{i}_j\}_{j=1}^Q$ is a set of collocation points uniformly sampled from $[0,1]$. In this example, we set $Q = 100$ and sample $N=5 \times 10^4$ different  initial conditions from a uniform distribution $\mathcal{U}(-3, 3)$. We train the physics-informed DeepONet for $3 \times 10^5$ iterations of gradient descent using Adam optimizer, and apply Algorithm \ref{alg: long_time_integration} to the same initial condition as in the example presented in Figure  \ref{fig: PINN_pendulum_s_pred}. As shown in Figure \ref{fig: PI_deeponet_pendulum_s_pred}, we observe that the model predictions are in good agreement with the exact solution. The resulting relative $L^2$ errors of $s^{(1)}$ and $s^{(2)}$ are $1.72\%$ and $1.63\%$, respectively. It is worth emphasising that the input function space $\mathcal{U}$ should be large enough to cover as many potential states of the underlying ODE/PDE system as possible. Otherwise, the trained model may not generalize very well for out-of-distribution initial conditions, possibly leading to large errors or even erroneous predictions.


\section{Results}\label{sec: results}

To demonstrate the effectiveness of  the proposed algorithm , we provide a series of comprehensive  numerical studies for solving various long-time integration problems. Throughout all benchmarks, we will employ a modified fully-connected neural network architecture \cite{wang2020understanding} which has been empirically proved  to outperform the standard fully-connected neural network architectures \cite{wang2020understanding, wang2021learning, hennigh2020nvidia}. The forward pass is defined by
\begin{align}
    &U = \phi(X W^1 + b^1), \ \  V = \phi(X W^2 + b^2) \label{eq:ADGM_1}\\
    &H^{(1)} = \phi(X W^{z,1} + b^{z, 1}) \\
    &Z^{(k)} = \phi(H^{(k)}W^{z,k} + b^{z, k}), \ \ k=1, \dots, L \\
    &H^{(k+1)} = (1 - Z^{(k)}) \odot U  +  Z^{(k)}  \odot V, \ \  k=1, \dots, L \\
   & f_{\bm{\theta}}(x) = H^{(L+1)}W  + b \label{eq:ADGM_2},
\end{align}
where $X$ denotes the network inputs, and $\odot$ denotes element-wise multiplication. The parameters of this model are essentially the same as in  conventional fully-connected architectures, with the addition of the weights and biases used by the two transformer sub-networks, i.e.,
\begin{align}
    \theta = \{W^1, b^1, W^2, b^2, (W^{z,l}, b^{z,l})_{l=1}^L, W,b  \}
\end{align}
In all cases we employ hyperbolic tangent activation functions (Tanh) and initialize all trainable parameters using the Glorot normal scheme \cite{glorot2010understanding}. Physics-informed DeepONet models are trained via mini-batch gradient descent with a batch-size of $10,000$ using the Adam optimizer with default settings \cite{kingma2014adam}, and an exponential learning rate decay with a decay-rate of 0.9 every 5,000 iterations. In this work, we tuned these hyper-parameters manually, without attempting to find the absolute best hyper-parameter setting. This process can be automated in the future leveraging effective techniques for meta-learning and hyper-parameter optimization \cite{finn2017model}. Additional details related to performance metrics, computational cost, hyper-parameters and training details are discussed in the Appendix. All results presented in this section can be reproduced using open-source code that will be made publicly available at \url{https://github.com/PredictiveIntelligenceLab/Long-time-Integration-PI-DeepONets}.


\subsection{Inhomogeneous ODEs}

As our first example, we start with a simple benchmark to illustrate how to generalize Algorithm \ref{alg: long_time_integration} to inhomogeneous differential equations. Particularly, we consider a 1D ODE of the form
\begin{align}
    \label{eq: ODE_sin}
    &\frac{d s}{d t} = \cos(t), \quad t \in  [0, T],\\
    \label{eq: ODE_sin_IC}
    &s(0) = 0.
\end{align}
The objective is to use Algorithm \ref{alg: long_time_integration} to learn the ODE solution for $T=10^3$. 
However, one may note that the proposed algorithm cannot be directly applied to this problem because of the forcing term. For example, assume that $G_{\bm{\theta}}$ is a trained physics-informed DeepONet that approximates the solution operator from the initial condition to the associated solution of the ODE (\ref{eq: ODE_sin}) for $t \in [0, 1]$ and  $\Tilde{s}(t)$ is an inferred solution corresponding to some initial condition. To obtain the solution in $[1,2]$, the trained DeepONet is required to yield the predicted solution governed by the following ODE
\begin{align*}
       &\frac{d s}{d t} = \cos(t+1), \quad t \in  [0, 1],\\
    &s(0) = \Tilde{s}(1),
\end{align*}
which is impossible because the trained model would be designing to work with a fixed forcing term. Fortunately, one can easily overcome this technical difficulty by solving the following parametric ODE
\begin{align}
    &\frac{d s}{d t} = u(t), \quad t \in [0, \Delta t]\\
    & s(0) = u_0
\end{align}
where both the initial condition and the forcing term can be considered as random inputs.

To represent the solution map of the above parametric ODE with a DeepONet $G_{\bm{\theta}}$, we  employ a single trunk network for extracting latent representations of input coordinates, but two  separate branch networks for representing the initial condition and the forcing term, respectively. The final output of the modified DeepONet architecture can be then obtained as
\begin{align}
    \label{eq: modified_deeponet_output}
    G_{\bm{\theta}}(\bm{u}, u_0)(t) = \sum_{k=1}^{q} \underbrace{b_{k}^{(1)}\left(u\left(t_{1}\right), u\left(t_{2}\right), \ldots, u\left(t_{m}\right)\right)}_{\text {branch I}} \cdot
    \underbrace{b_{k}^{(2)}\left(u_0\right)}_{\text {branch II}} \cdot
    \underbrace{t_{k}(t)}_{\text {trunk }},
\end{align}
where $\{b_k^{(1)}\}_{k=1}^q, \{b_k^{(2)}\}_{k=1}^q$ are outputs of two branch networks, respectively, and $u_0$ denotes the input initial condition. Also, $\bm{u} = [u(t_1), u(t_2), \dots, u(t_m)]$ represents the input forcing term evaluated at a set of fixed sensors $\{t_k\}_{k=1}^m \subset [0, \Delta t]$. In this example, we take $\Delta t = 1$, $m=100$ and then $\{t_i\}_{i=1}^m$ are equi-spaced grid points in $[0,1]$. Besides, all neural networks are 7-layer modified fully-connected neural networks (see equations (\ref{eq:ADGM_1})-(\ref{eq:ADGM_2})) with $100$ units per hidden layer. The resulting  physics-informed DeepONet can be trained by minimizing the following loss 
\begin{align}
        \mathcal{L}(\bm{\theta}) = \mathcal{L}_{\text{ic}}(\bm{\theta}) + \mathcal{L}_{r}(\bm{\theta}),
\end{align}
where
\begin{align}
     &\mathcal{L}_{\text{ic}}(\bm{\theta}) = \frac{1}{N}\sum_{i=1}^N   \left|G_{\bm{\theta}}(\bm{u}^{(i)}, u_0^{(i)})( 0 ) - u_0^{(i)}\right|^2, \\
     & \mathcal{L}_{r}(\bm{\theta}) =  \frac{1}{NQ}\sum_{i=1}^N  \sum_{j=1}^Q  \left|G_{\bm{\theta}}(\bm{u}^{(i)}, u_0^{(i)})( t_j^{(i)}) - u^{(i)}(t_j^{(i)}) \right|^2.
\end{align}
We generate "training data" by randomly sampling $N = 5 \times 10^4$ inputs $\{\bm{u}^{(i)}\}_{i=1}^N, \{u_0^{(i)}\}_{i=1}^N$ from a Gaussian random field (GRF)  with a length scale of $l=0.5$ and a uniform distribution $\mathcal{U}(-2, 2)$, respectively. For each paired input sample $(\bm{u}^{(i)}, u^{(i)}_0)$, the collocation points $\{t_j^{(i)}\}_{j=1}^Q$ are uniformly sampled in the unit interval with $Q = 100$. 

\begin{figure}
    \centering
    \includegraphics[width=0.9\textwidth]{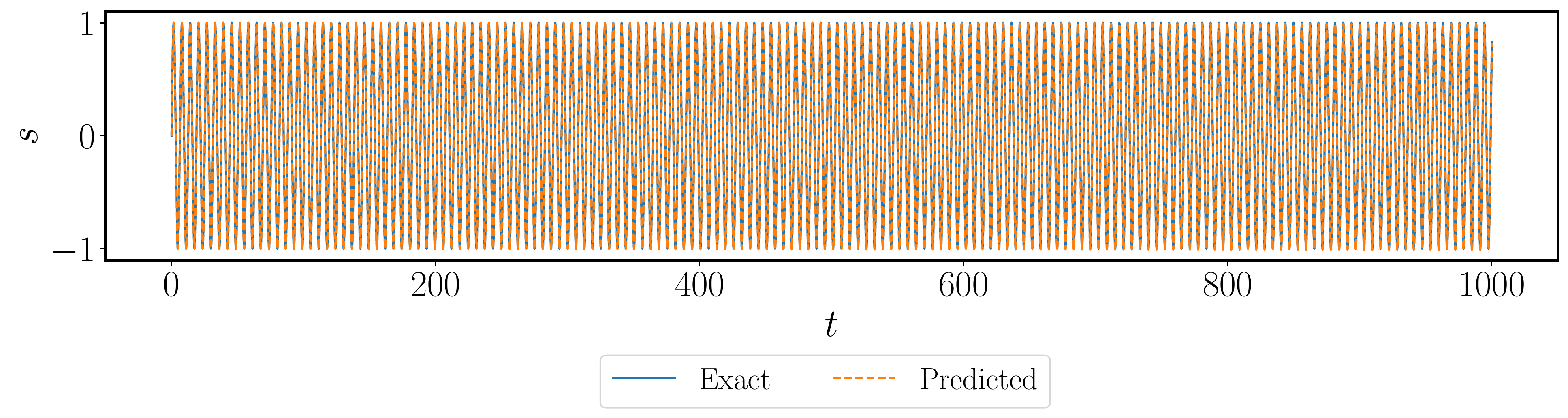}
     \caption{{\em Inhomogeneous ODE:} Exact solution versus the  predicted solution of a trained physics-informed DeepONet using Algorithm \ref{alg: long_time_integration} to integrate the ODE system of equations (\ref{eq: ODE_sin})  - (\ref{eq: ODE_sin_IC}) up to $T = 1000$. The relative $L^2$ error is  $0.84\%$.}
    \label{fig: PI_deeponet_ODE_s_pred}
\end{figure}

\begin{figure}
    \centering
    \includegraphics[width=0.4\textwidth]{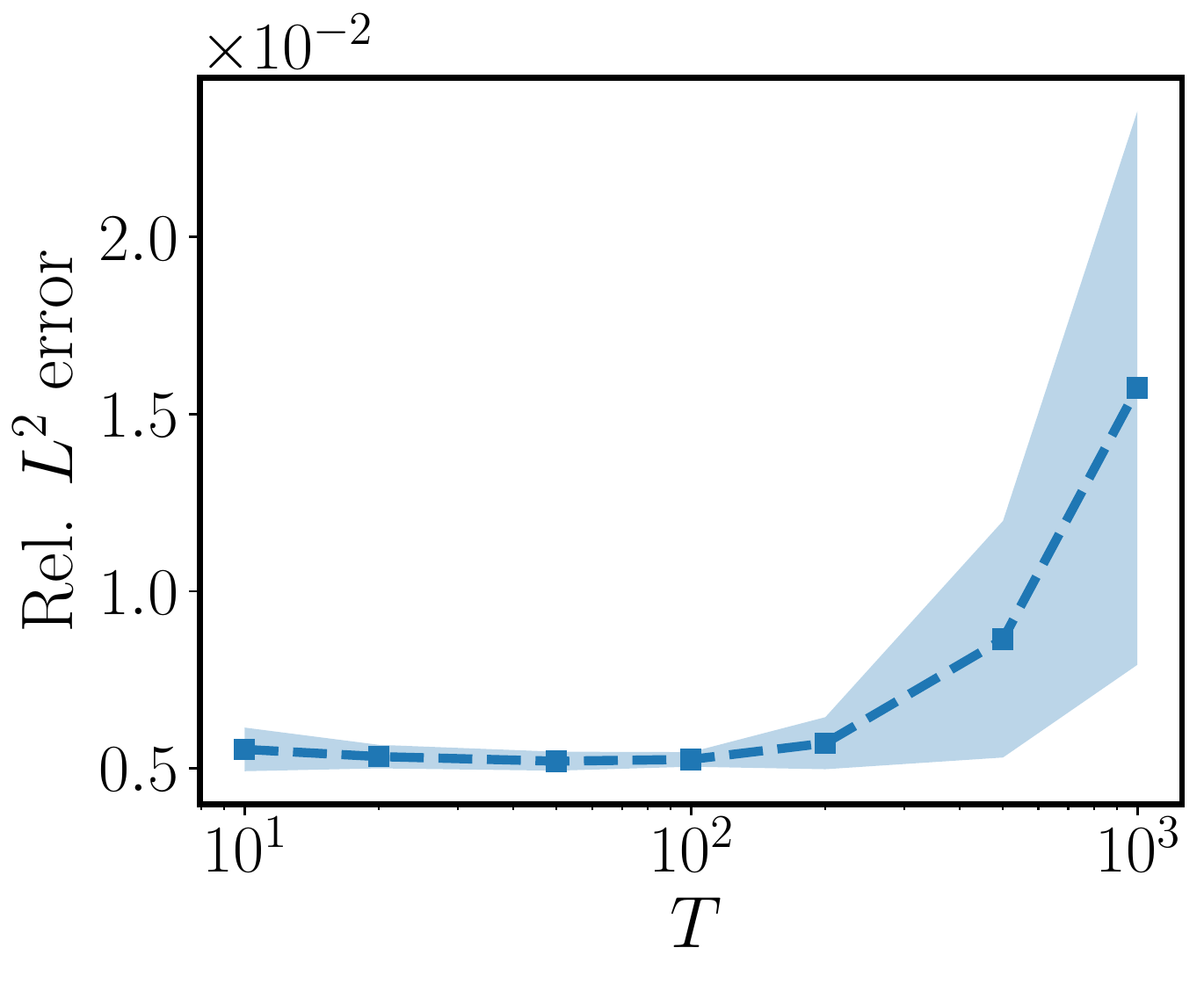}
    \caption{{\em  linear ODE:} Relative $L^2$ prediction error of a physics-informed DeepONet as a function of the final time $T$, averaged over 100 different examples in the test data-set.}
    \label{fig: PI_deeponet_ODE_error}
\end{figure}

Figure \ref{fig: PI_deeponet_ODE_s_pred} shows the inferred solution of the ODE system (\ref{eq: ODE_sin})-(\ref{eq: ODE_sin_IC}) obtained by applying Algorithm \ref{alg: long_time_integration} to a physics-informed DeepONet that was trained for $2 \times 10^5$ gradient descent iterations via the Adam optimizer \cite{kingma2014adam}. It can be observed that the model prediction achieves an excellent agreement with the exact solution as the resulting relative $L^2$ error over the whole time-domain is $0.84\%$. More interestingly, Algorithm \ref{alg: long_time_integration} also performs very well for different initial conditions sampled from a different function space than the one used to train the model ($\mathcal{U}(-3, 3)$). To illustrate this, we randomly sample $N =100$ different $u_0$ from a uniform distribution $\mathcal{U}(-\frac{1}{2}, \frac{1}{2})$ and obtain the inferred solution using the proposed algorithm. 
The relative $L^2$ error at the final time $T$ averaged across all examples in the test data-set is displayed in Figure \ref{fig: PI_deeponet_ODE_error}. One can see that all the inferred solutions keep almost the same accuracy ($\sim 0.5\%$) up to $T = 100$, while the approximation errors tend to accumulates as time $T$ increases. Additional visualizations for different input samples are provided in Appendix Figure \ref{fig: PI_deeponet_ODE_s_pred_examples}. 

\subsection{Stiff chemical kinetics}


Our next example aims to demonstrate the capability of the proposed algorithm to perform long-time integration of stiff ODEs, which are generally hard to solve by conventional PINNs \cite{ji2020stiff}, as well as classical numerical methods \cite{wanner1996solving, iserles2009first}. To this end, we consider a classical stiff chemical kinetics problem describing the kinetics of an autocatalytic reaction \cite{robertson1966solution, wanner1996solving}
\begin{align}
     \label{eq: stiff_ode}
    &\frac{d s_{1}}{d t}=-k_{1} s_{1} +k_{3} s_{2} s_{3}, \\
    & \frac{d s_{2}}{d t}=k_{1} s_{1}-k_{2} s_{2}^{2}-k_{3} s_{2} s_{3}, \\
    &\frac{d s_{3}}{d t}=k_{2} s_{2}^{2},
\end{align}
with the initial condition
\begin{align}
    \label{eq: stiff_ode_ic}
    s_1(0) = 1, \quad s_2(0) = s_3(0) = 0,
\end{align}
where $s_1, s_2, s_3$ denote the concentrations of different chemical species, while the reaction rate constants are $k_{1}=0.04, k_{2}=3 \times 10^{4}, k_{3}=10^{7}$. It worth pointing out that the different species have very different reaction timescales, especially for $s_2$, which consequently results in a very stiff system.

We proceed by employing a physics-informed DeepONet $\bm{G}_{\bm{\theta}}$ to represent the solution operator $G$ mapping the initial condition $\bm{u}\in \R^3$ to the solution of the kinetic system in time interval $[0, 1]$ (i.e., $\Delta t = 1$).  Since the ODE system has three variables,  the forward pass of the DeepONet is given by equation (\ref{eq: deeponet_multiple_outputs}) where we take $i=3$ and $[q_0, q_1, q_2, q_3] = [0, 100, 200, 300]$. Accordingly, the ODE residuals are defined by
\begin{align}
    \label{eq: PI_deeponet_stiff_ode_res_1}
    &\mathcal{R}_{\bm{\theta}}^{(1)}[\bm{u}](t) = \frac{d G^{(1)}_{\bm{\theta}}(\bm{u})(t)}{ dt} +  k_1 G^{(1)}_{\bm{\theta}}(\bm{u})(t)  - k_3 G^{(2)}_{\bm{\theta}}(\bm{u})(t) G^{(3)}_{\bm{\theta}}(\bm{u})(t) , \\
    &\mathcal{R}_{\bm{\theta}}^{(2)}[\bm{u}](t) =  \frac{d G^{(2)}_{\bm{\theta}}(\bm{u})(t)}{ dt} -  k_1 G^{(1)}_{\bm{\theta}}(\bm{u})(t)  + k_2  [G^{(2)}_{\bm{\theta}}(\bm{u})(t)]^2 + k_3 G^{(2)}_{\bm{\theta}}(\bm{u})(t) G^{(3)}_{\bm{\theta}}(\bm{u})(t), \\
      &\mathcal{R}_{\bm{\theta}}^{(3)}[\bm{u}](t) = \frac{d G^{(3)}_{\bm{\theta}}(\bm{u})(t)}{ dt}  - k_2  [G^{(2)}_{\bm{\theta}}(\bm{u})(t)]^2,
\end{align}
where $\bm{G}_{\bm{\theta}} = [G^{(1)}_{\bm{\theta}}, G^{(2)}_{\bm{\theta}}, G^{(3)}_{\bm{\theta}}]$ represents the three species $[s_1, s_2, s_3]$ in the kinetic system, and $\bm{u} = [u_1, u_2, u_3]$ denotes the initial concentration for each state variable. This allows us to formulate the physics-informed loss function
\begin{align}\label{eq:stiff_loss}
        \mathcal{L}(\bm{\theta}) = \mathcal{L}_{\text{ic}}(\bm{\theta}) + \mathcal{L}_{r}(\bm{\theta}),
\end{align}
where
\begin{align}
     &\mathcal{L}_{\text{ic}}(\bm{\theta}) = \sum_{k=1}^3 \lambda_k \mathcal{L}_{\text{ic}}^{(k)}(\bm{\theta}) =    \sum_{k=1}^3  \lambda_k \left[ \frac{1}{N}\sum_{i=1}^N \left|G_{\bm{\theta}}^{(k)}(\bm{u}^{(i)})( 0 ) - u_k^{(i)}\right|^2 \right] , \\
     & \mathcal{L}_{r}(\bm{\theta}) =  \sum_{k=1}^3 \mathcal{L}_{r}^{(k)}(\bm{\theta}) =\sum_{k=1}^3 \left[\frac{1}{NQ}\sum_{i=1}^N  \sum_{j=1}^Q  \left|\mathcal{R}_{\bm{\theta}}^{(k)}[\bm{u}^{(i)}](t^{(i)}_j)  \right|^2 \right].
\end{align}
In this example, the branch and trunk networks parametrizing $\bm{G}_{\bm{\theta}}$ are two 7-layer modified fully-connected neural networks with $100$ neurons per hidden layer (see equations (\ref{eq:ADGM_1})-(\ref{eq:ADGM_2})). The model is trained on a data-set created by randomly sampling $N = 5 \times 10^4$ initial states $\{\bm{u}^{(i)}\}_{i=1}^N = \{[u_1^{(i)}, u_2^{(i)}, u_3^{(i)}]\}_{i=1}^N$, where $u_1^{(i)}, u_3^{(i)} \sim \mathcal{U}(0, 1)$,  and $u^{(i)}_2 \sim \mathcal{U}(0, 10^{-4})$, for all $i$. For each input sample $\bm{u}^{i}$ we have $Q = 10^3$, and $\{t^{(i)}_j\}_{j=1}^Q$ are collocation points sampled from $\mathcal{U}(0, 1)$. Furthermore, because the species $s_2$ is typically about four order of magnitudes smaller than $s_1$ and $s_3$, we manually re-scale the range of $G_{\bm{\theta}}^{(2)}$ by multiplying it by $10^{-4}$, and set $[\lambda_1, \lambda_2, \lambda_3] = [1, 10^6, 1]$ to the corresponding imbalance between the different loss functions. To generate the test data-set, we randomly sample $100$ initial conditions, and obtain the corresponding numerical solutions in $[0, 500]$ by integrating the ODE using the Radau scheme \cite{hairer1999stiff}.

\begin{figure}
     \centering
     \begin{subfigure}[b]{0.6\textwidth}
         \centering
    \includegraphics[width=\textwidth]{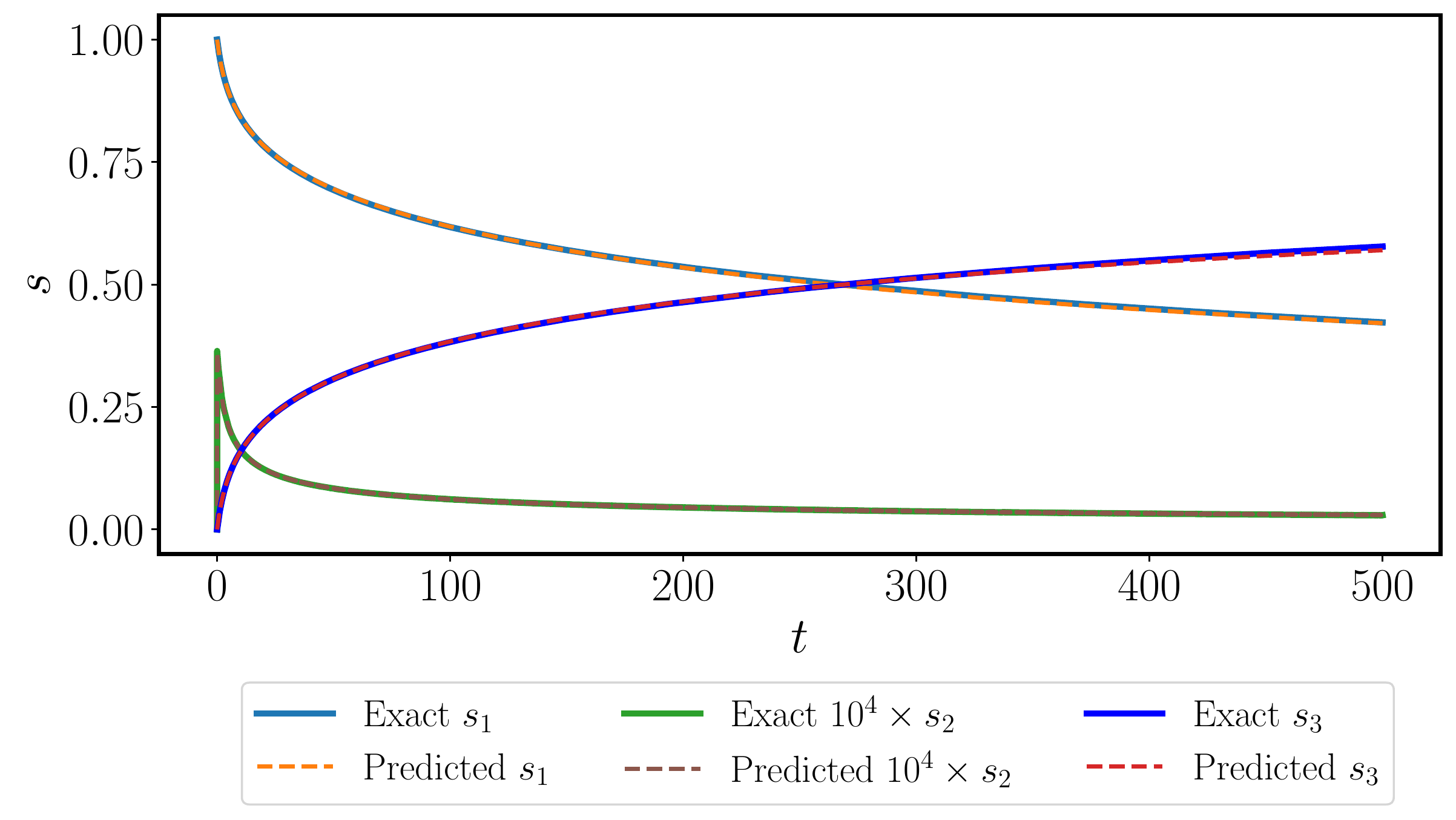}
     \end{subfigure}
     \begin{subfigure}[b]{0.6\textwidth}
         \centering
    \includegraphics[width=\textwidth]{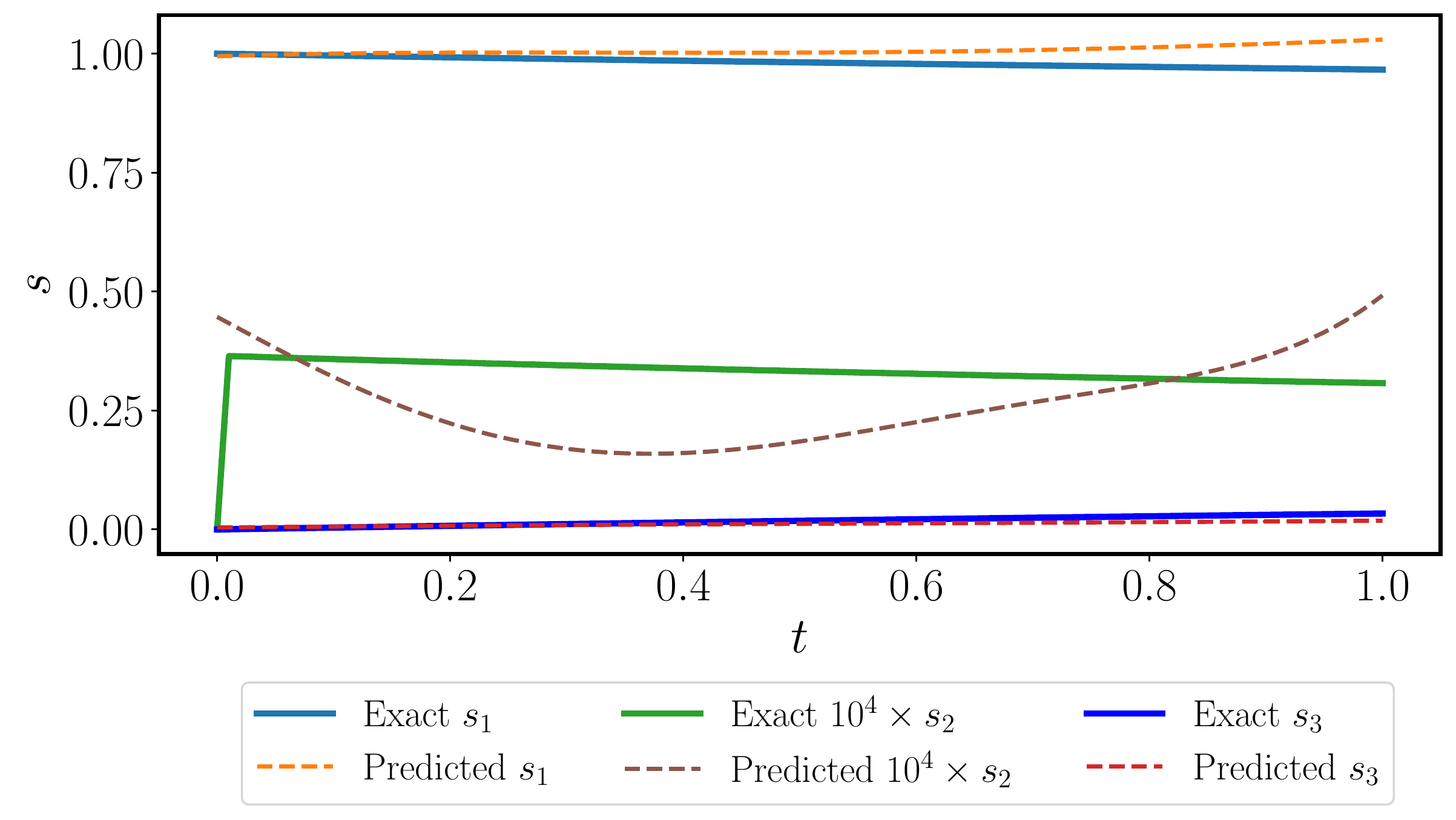}
     \end{subfigure}
        \caption{{\em Stiff chemical kinetics:} {\em Top:} Reference solution versus the prediction of a trained scaled physics-informed DeepONet using Algorithm \ref{alg: long_time_integration} for integrating the ODE system of equations (\ref{eq: stiff_ode}) - (\ref{eq: stiff_ode_ic}) up to $T = 500$. The relative $L^2$ errors for $s_1, s_2, s_3$ are $0.38\%,  0.56\%, 0.64\%$, respectively. {\em Bottom:} Reference solution versus the prediction of a trained un-scaled physics-informed DeepONet.  The relative $L^2$ errors for $s_1, s_2, s_3$ are $3.05\%,  37.01\%, 41.93\%$, respectively. We can observe that the trained model fails to yield accurate predictions even for a short-time interval corresponding to $T=1$.}
        \label{fig: PI_deeponet_stiff_ODE_s_preds}
\end{figure}

\begin{figure}
     \centering
     \begin{subfigure}[b]{0.4\textwidth}
       \centering
    \includegraphics[width=1.0\textwidth]{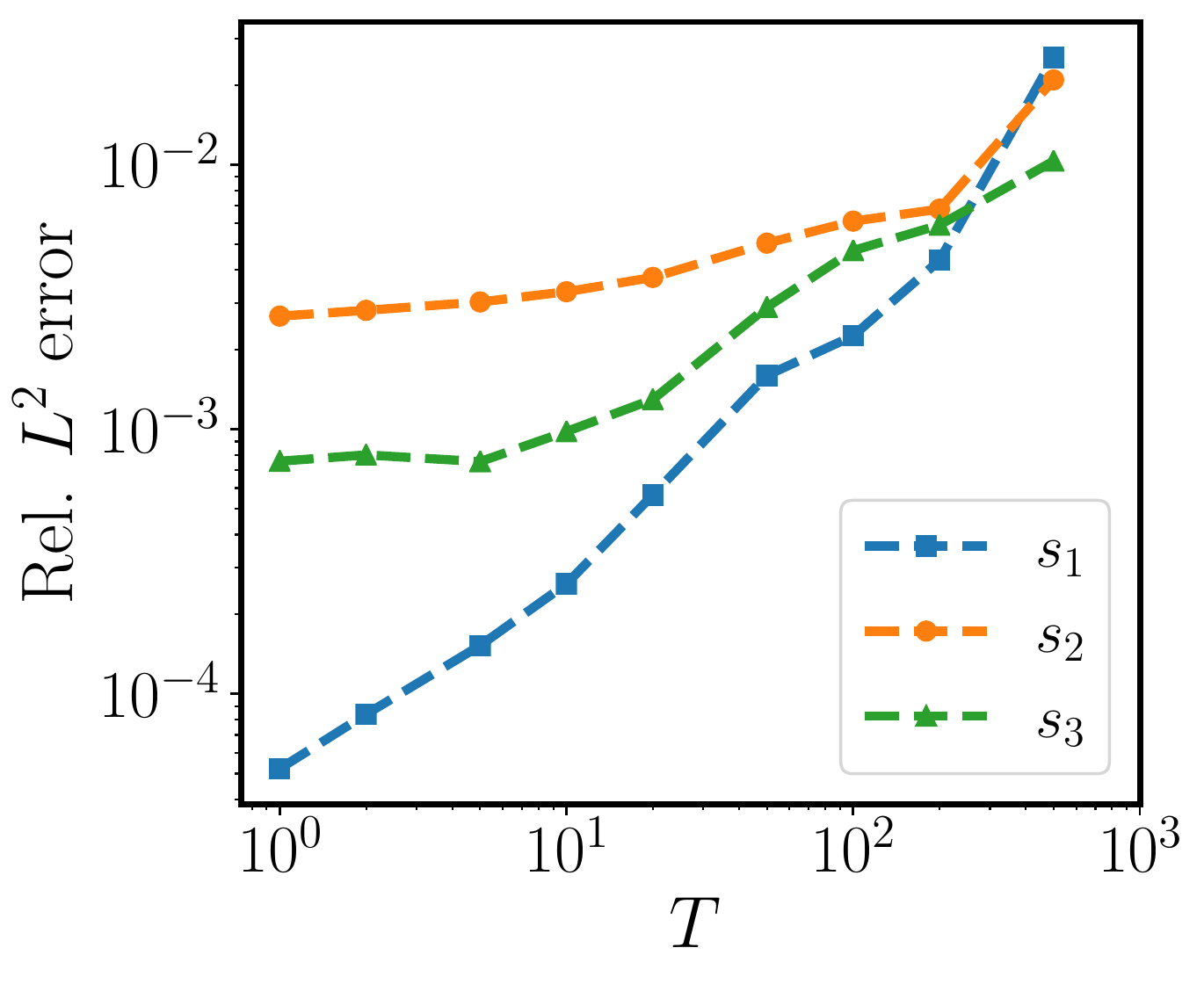}
     \end{subfigure}
     \begin{subfigure}[b]{0.4\textwidth}
         \centering
    \includegraphics[width=1.0\textwidth]{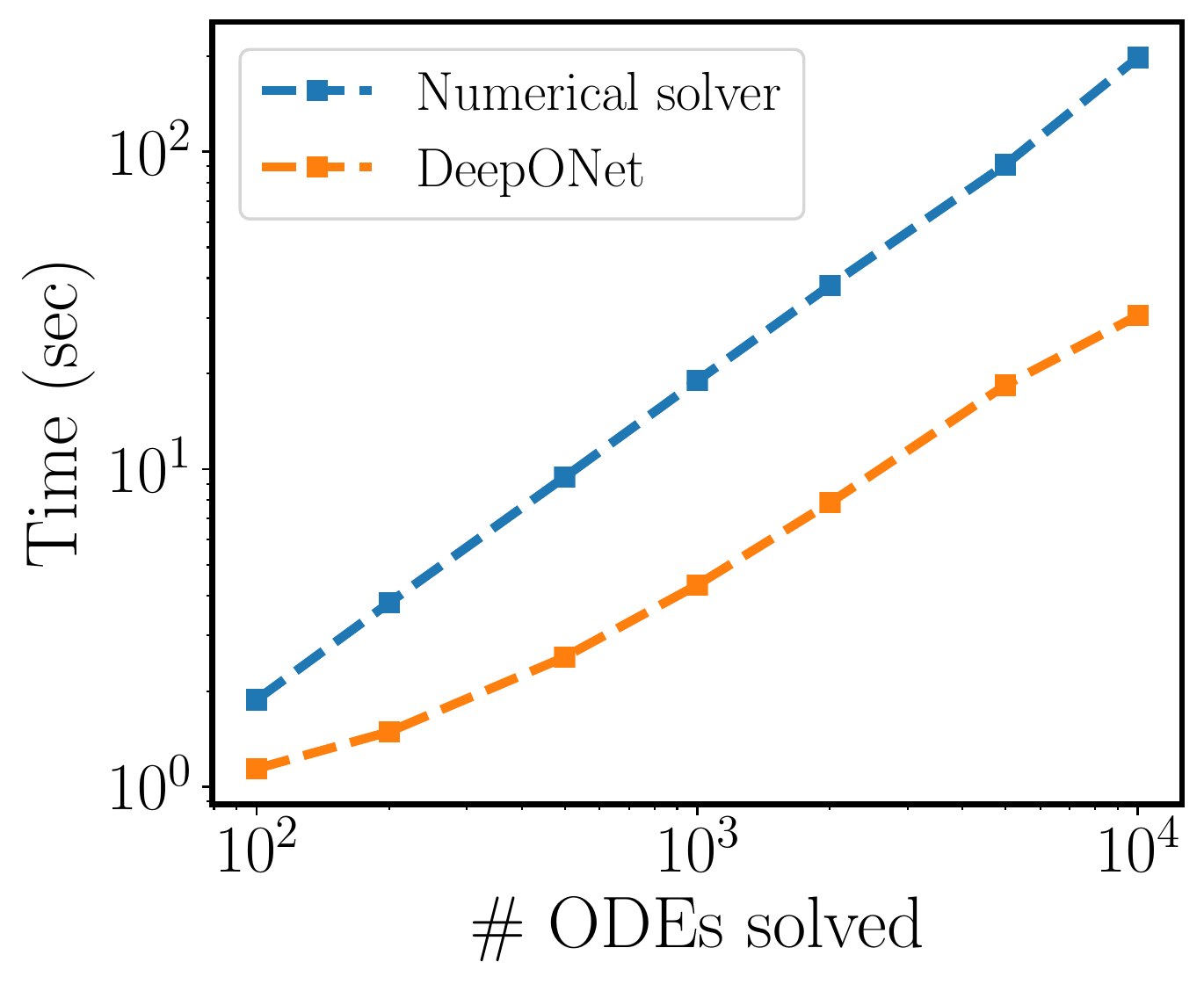}
     \end{subfigure}
        \caption{{\em Stiff chemical kinetics:} {\em Left:} Relative $L^2$  prediction error of a trained physics-informed DeepONet as a function of the final prediction time $T$, averaged over 100 different examples in the test data-set. {\em Right:}  Computational cost (sec) for performing inference with a trained physics-informed DeepONet model versus the time taken for solving the ODE with a conventional Radau scheme \cite{hairer1999stiff}.}
        \label{fig: PI_deeponet_stiff_ODE_s_error_time}
\end{figure}

We train the physics-informed DeepONet by minimizing the loss of equation (\ref{eq:stiff_loss}) for $4 \times 10^5$ gradient descent iterations using the Adam optimizer. The inferred solution obtained using the trained model subject to the initial condition (\ref{eq: stiff_ode_ic}) is presented in the top panel of Figure \ref{fig: PI_deeponet_stiff_ODE_s_preds}, from which we can observe an excellent agreement between the model inference and the ground truth with  relative $L^2$ errors of $0.38\%,  0.56\%, 0.64\%$ for $s_1, s_2, s_3$, respectively. Some representative visualizations for different initial conditions are shown in Appendix Figure \ref{fig: PI_deeponet_stiff_ODE_s_pred_examples}. 
Moreover, the relative $L^2$ error of the model predictions, averaged over all 100 examples in the test data-set, is visualized in Figure \ref{fig: PI_deeponet_stiff_ODE_s_error_time}. From these figures, one may conclude that the trained model is capable of yielding accurate long-time predictions for different initial conditions. 

To demonstrate the necessity of assigning weights to the different terms in the physics-informed loss function, we train the same model without any weights under exactly the same hyper-parameter settings,  and the result is summarized in the bottom panel of Figure \ref{fig: PI_deeponet_stiff_ODE_s_preds}. It is evident that the un-scaled physics-informed DeepONet fails to learn the correct solution even on a unit interval, let alone long-time integration. These observations are consistent with the findings reported in \cite{wang2020understanding, wang2020and, wang2021learning}, highly suggesting that rescaling the network outputs and the loss functions is a prerequisite for achieving good predictive accuracy, especially for problems that exhibit stiff and multi-scale behavior. Although here we manually assign weights assuming some prior knowledge of the underlying ODE model form, it must be emphasized that such a manual approach is time-consuming or even impractical for many realistic scenarios involving high-dimensional state spaces, multi-physics, complicated loss functions, etc. Hence, we point out that there is an urgent need to understand the training dynamics of physics-informed DeepONets and develop effective training algorithms that can automatically balance the interplay between different terms in the corresponding loss functions. 

Once trained, the model can be rapidly queried to return prediction at short-time intervals. Each query typically takes $\mathcal{O}(10^{-3}\text{ms})$  on a single Nvidia V100 GPU. Constructing the global PDE/ODE solution in large temporal domains using Algorithm \ref{alg: long_time_integration} requires $N\sim\mathcal{O}(10)- \mathcal{O}(10^3)$ evaluations of the trained model, typically leading to a total inference time of $\sim \mathcal{O}(10^{-2})$ sec for a given initial condition. Multiple initial conditions can also be simulated at once, as our JAX \cite{jax2018github} implementation is trivial vectorized and parallelized on GPU hardware. As demonstrated in Figure \ref{fig: PI_deeponet_stiff_ODE_s_error_time}, the proposed framework can perform long-time integration of more than 1,000 initial conditions in $\mathcal{O}(1)$ second, yielding a $\sim$10x-50x speedup compared to a traditional numerical solver. Unlike traditional numerical solvers that are heavily specialized to a specific type of dynamic behavior, this cost remains nearly constant for all examples considered in this work, regardless of the ODE/PDE system that is simulated, as it merely amounts to the cost associated with evaluating the forward pass of the trained DeepONet model.

\subsection{Wave propagation}

\begin{figure}
    \centering
    \includegraphics[width=0.7\textwidth]{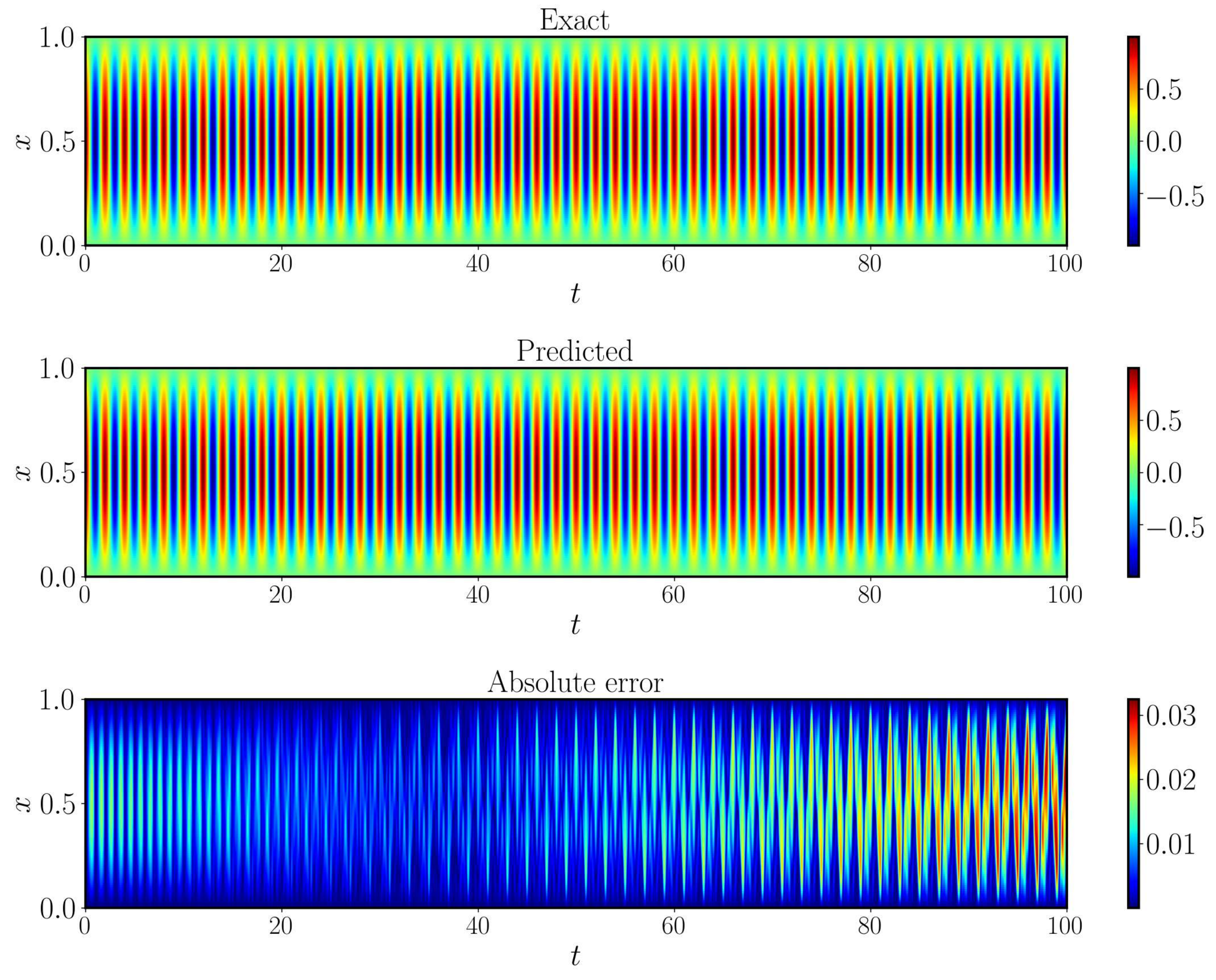}
    \caption{{\em Wave propagation:} Comparison between the exact and the predicted solution of a trained physics-informed DeepONet using Algorithm \ref{alg: long_time_integration} for integrating the PDE (\ref{eq: wave_pde})  - (\ref{eq: wave_ic_2}) up to $T = 100$. The relative $L^2$ error is $1.57\%$. }
    \label{fig: PI_deeponet_wave_s_pred}
\end{figure}

In the numerical examples presented so far we mainly focus on ODEs.
To highlight the ability of the proposed algorithm to handle long-time integration problems for PDEs, we begin by considering the 1D wave equation 
\begin{align}
    \label{eq: wave_pde}
    &\frac{\partial^2 s}{\partial t^2} = c^2 \frac{\partial^2 s }{\partial x^2} , \quad (x, t) \in [0,1] \times [0, T], \\
    & s(0, t) = s(1, t) = 0, \quad t \in [0, T],\\
    &s(x, 0) = \sin( \pi x), \quad x \in [0, 1],\\
    \label{eq: wave_ic_2}
    &\frac{\partial s}{\partial t}(x, 0) = 0,
\end{align}
where we take $c=1$ and $T = 100$.  The exact solution is given by
\begin{align}
    s(x,t) = \sin(\pi x) \cos(c \pi t).
\end{align}
To solve this problem, we parametrize the initial condition by a Gaussian random field with a length scale $l = 0.5$, and use a DeepONet $G_{\bm{\theta}}$ to represent the solution map from initial conditions to the associated PDE solutions. Here, the branch and trunk networks are  5-layer modified fully-connected neural networks (see equations (\ref{eq:ADGM_1})-(\ref{eq:ADGM_2})) with 200 units per hidden layer. The parameters of the physics-informed DeepONet can be trained by minimizing the following loss function
\begin{align}
        \mathcal{L}(\bm{\theta}) = \mathcal{L}_{\text{ic}}(\bm{\theta}) + \mathcal{L}_{\text{bc}}(\bm{\theta}) + \mathcal{L}_{r}(\bm{\theta}),
\end{align}
where
\begin{align}
     &\mathcal{L}_{\text{ic}}(\bm{\theta}) = \frac{1}{NP} \sum_{i=1}^N \sum_{j=1}^P \left[ 
     \left|G_{\bm{\theta}}(\bm{u}^{(i)}) (x_{\text{ic}, j}^{(i)}, 0) - u^{(i)}(x_{\text{ic}, j}^{(i)})\right|^2 + \left|  \frac{\partial G_{\bm{\theta}}(\bm{u}^{(i)})}{\partial t}  (x_{\text{ic}, j}^{(i)}, 0) \right|^2
     \right]     , \\
      &\mathcal{L}_{\text{bc}}(\bm{\theta}) = \frac{1}{NP} \sum_{i=1}^N  \sum_{j=1}^P \left[ \left|G_{\bm{\theta}}(\bm{u}^{(i)}) (0, t_{\text{bc}, j}^{(i)})\right|^2 + \left|G_{\bm{\theta}}(\bm{u}^{(i)}) (1, t_{\text{bc}, j}^{(i)})\right|^2 \right]   , \\
     & \mathcal{L}_{r}(\bm{\theta}) =  \frac{1}{NQ}\sum_{i=1}^N  \sum_{j=1}^Q  \left|\frac{\partial^2 G_{\bm{\theta}}(\bm{u}^{(i)}) }{\partial t^2} (x_{r, j}^{(i)}, t_{\text{r}, j}^{(i)}) - c^2
     \frac{\partial^2 G_{\bm{\theta}}(\bm{u}^{(i)}) }{\partial x^2} (x_{r, j}^{(i)}, t_{\text{r}, j}^{(i)})
     \right|^2.
\end{align}
Here, we take $\Delta t=1$, $N=10^4$, $m=P =100$ and $Q =200$. In particular,  for every input sample $\bm{u}^{(i)}$, $\{x_{\text{ic}, j}^{(i)}\}_{j=1}^P, \{t_{\text{bc}, j}^{(i)}\}_{j=1}^P$ and $\{x_{r, j}^{(i)}, t_{\text{r}, j}^{(i)}\}_{j=1}^Q$ are uniformly sampled in the computational domain $[0, 1] \times [0, 1]$  for enforcing the initial/boundary conditions and the PDE residual, respectively.  We train the physics-informed DeepONet for $2 \times 10^5$ iterations of gradient descent using the Adam optimizer. A comparison of the long-time prediction  against the ground truth is shown in Figure \ref{fig: PI_deeponet_wave_s_pred}. We can observe a good agreement between the predicted and the exact solution, yielding a relative error of $1.57e-02$ in the relative $L^2$ norm.

\subsection{Diffusion-reaction dynamics}

Next,  we present a study on the effect of $\Delta t$ in Algorithm \ref{alg: long_time_integration} by solving a long-time integration problem involving  non-linear diffusion-reaction dynamics, as described by the following PDE system
\begin{align}
      &\frac{\partial s}{\partial t}=D \frac{\partial^{2} s}{\partial x^{2}}+k s^{2}, \quad (x,t) \in (0,1] \times (0, T], \\
      & s(x, 0) = u(x), \quad x \in [0, 1]\\
      & s(0, t) = s(1, t) = 0, \quad t \in [0, T],
\end{align}
where $T = 50$,  $D = 0.001$ is the diffusion coefficient, and $k=  0.001$ is the reaction rate.  According to Algorithm \ref{alg: long_time_integration}, it suffices to learn the solution operator that maps the initial condition to the associated PDE solution in $[0, \Delta t]$. To this end, 
we approximate the operator by a DeepONet  $G_{\bm{\theta}}$ where the branch net and the trunk network are 5-layer modified fully-connected neural network (see equations (\ref{eq:ADGM_1})-(\ref{eq:ADGM_2})) with 100 units per hidden layer.  The physics-informed loss function is given by
\begin{align}
        \mathcal{L}(\bm{\theta}) = \mathcal{L}_{\text{ic}}(\bm{\theta}) + \mathcal{L}_{\text{bc}}(\bm{\theta}) + \mathcal{L}_{r}(\bm{\theta}),
\end{align}
where 
\begin{align}
     &\mathcal{L}_{\text{ic}}(\bm{\theta}) = \frac{1}{NP} \sum_{i=1}^N \sum_{j=1}^P  \left|G_{\bm{\theta}}(\bm{u}^{(i)}) (x_{\text{ic}, j}^{(i)}, 0) - u^{(i)}(x_{\text{ic}, j}^{(i)})\right|^2, \\
      &\mathcal{L}_{\text{bc}}(\bm{\theta}) = \frac{1}{NP} \sum_{i=1}^N  \sum_{j=1}^P \left[ \left|G_{\bm{\theta}}(\bm{u}^{(i)}) (0, t_{\text{bc}, j}^{(i)})\right|^2 + \left|G_{\bm{\theta}}(\bm{u}^{(i)}) (1, t_{\text{bc}, j}^{(i)})\right|^2 \right]   , \\
     & \mathcal{L}_{r}(\bm{\theta}) =  \frac{1}{NQ} \sum_{i=1}^N  \sum_{j=1}^Q \left|  \frac{\partial G_{\bm{\theta}}(\bm{u}^{(i)}) }{\partial t} (x_{r, j}^{(i)}, t_{\text{r}, j}^{(i)}) - D \frac{\partial^2 G_{\bm{\theta}}(\bm{u}^{(i)}) }{\partial x^2} (x_{r, j}^{(i)}, t_{\text{r}, j}^{(i)}) - k G_{\bm{\theta}}^2(\bm{u}^{(i)}) (x_{r, j}^{(i)}, t_{\text{r}, j}^{(i)}) \right|.
\end{align}
Here, $\bm{u}^{(i)}$ denotes initial conditions randomly sampled from a GRF with a length scale of $l=0.2$, and for each input sample, $\{x_{\text{ic}, j}^{(i)}\}_{j=1}^P, \{ t_{\text{bc}, j}^{(i)}\}_{j=1}^P$ and $\{x_{r, j}^{(i)}, t_{\text{r}, j}^{(i)}\}_{j=1}^Q$ are uniformly sampled in the computational domain $[0, 1] \times [0, \Delta t]$.
In this example, we take $N = 10^4$ and $m = P = Q = 100$.
To generate the test data-set, we  sample
$N = 100$ input functions $u(x)$ from the same GRF and solve the diffusion-reaction system in $[0, 1] \times [0, T]$  by a second-order implicit finite difference method.

\begin{figure}
    \centering
    \includegraphics[width=0.4\textwidth]{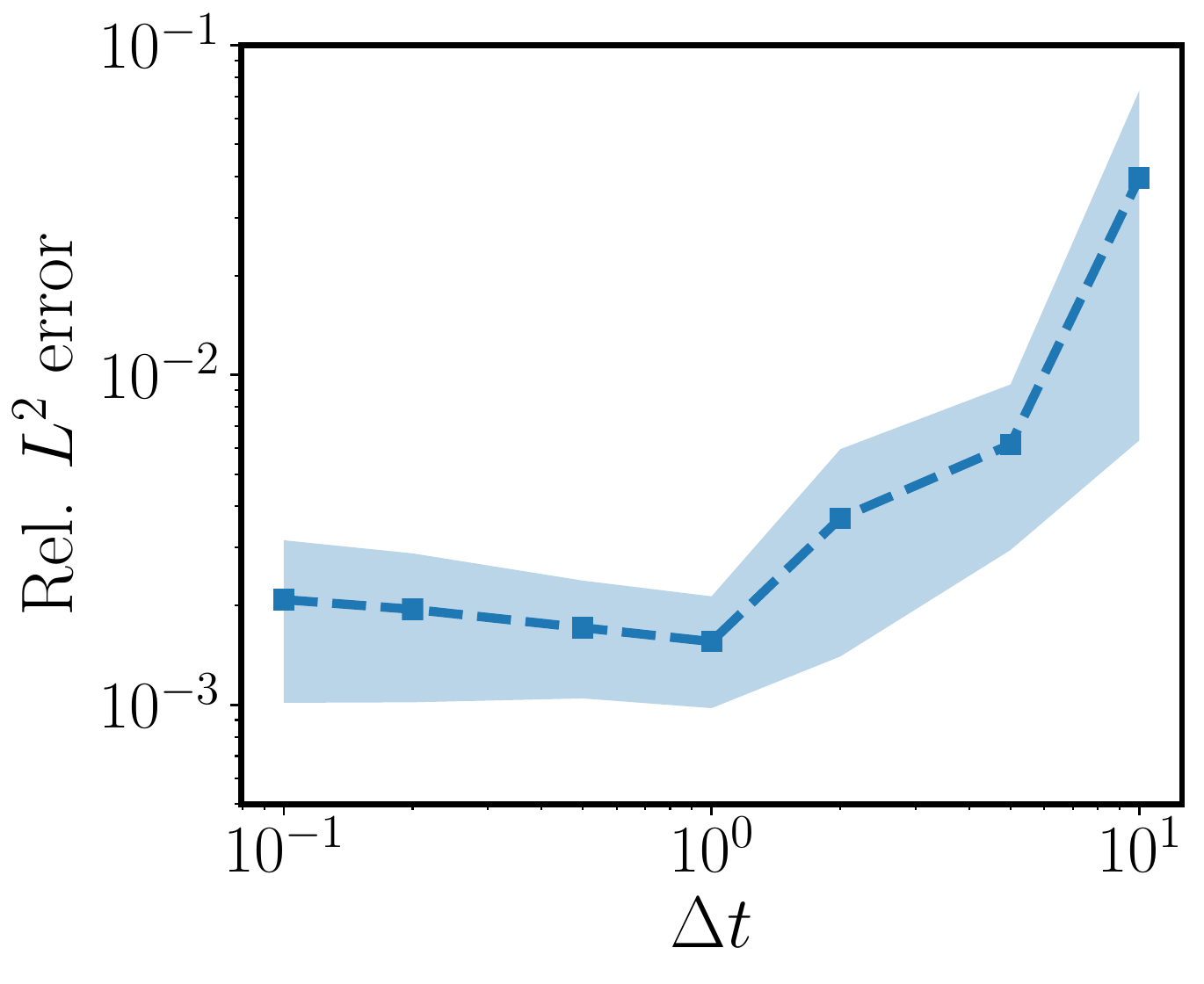}
    \caption{{\em Diffusion-reaction dynamics:}  Relative $L^2$ prediction errors of trained physics-informed DeepONets for different $\Delta t \in [0.1, 10]$, averaged over $100$ different examples in the test data-set.}
    \label{fig: PI_deeponet_DR_error}
\end{figure}

\begin{figure}
    \centering
    \includegraphics[width=0.8\textwidth]{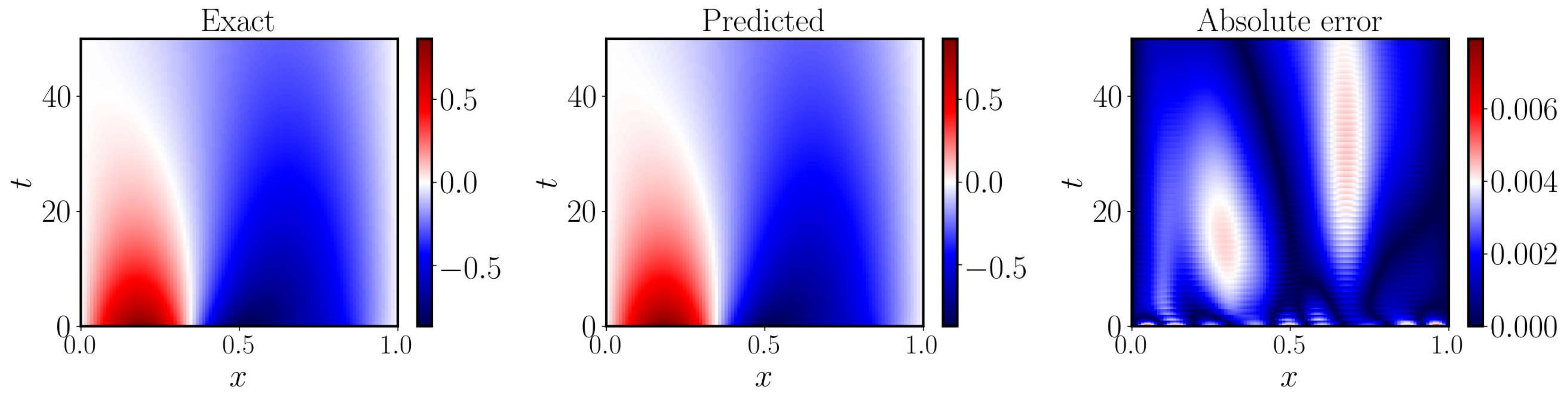}
    \caption{{\em Diffusion-reaction dynamics:} Reference solution versus the predicted solution of a trained physics-informed DeepONet using Algorithm \ref{alg: long_time_integration} for a representative input sample in the test data-set.  The relative $L^2$ prediction error is $0.67\%$.}
    \label{fig: Figures/DR/PI_deeponet_DR_s_pred.png}
\end{figure}

For different $\Delta t$, we train the physics-informed DeepONet for $2 \times 10^5$ iterations under exactly the same hyper-parameter settings. Figure \ref{fig: PI_deeponet_DR_error} shows the relative $L^2$ prediction errors of the trained models averaged over all examples in the test data-set. A key observation is that our approach is relatively robust against the choice of the time step size $\Delta t$, while trained model corresponding to $\Delta t = 1$ seems to yield the best predictive accuracy.
Moreover, the predicted solution of the best trained model for one representative input sample is shown in Figure \ref{fig: Figures/DR/PI_deeponet_DR_s_pred.png}. One can see that the predictions achieves an excellent agreement with the corresponding numerical estimations. The resulting relative $L^2$ error is $0.67\%$. Additional representative results corresponding to different initial conditions can be found in the Appendix Figure \ref{fig: PI_deeponet_DR_s_pred_examples}, which further verifies our conclusion.  

\subsection{Korteweg–De Vries equation}

In our last example we would like to emphasize the effectiveness of physics-informed DeepONets in integrating observational data and governing evolution equations in a data-efficient manner. To this end, we pursue the  long-time prediction of traveling solitons governed by the Korteweg–De Vries (KDV) equation, assuming that, instead of boundary conditions for the latent solution, only some sparse observations in a short-time interval are available. Specifically,  we set $\epsilon = 1.2 \times 10^{-1}$ and $\mu = 8 \times 10^{-4}$ and consider the following PDE system
\begin{align}
    \label{eq: KDV}
    &\frac{\partial s}{\partial t} + \epsilon  \frac{\partial s}{\partial x} s + \mu  \frac{\partial^3 s}{\partial x^3} = 0, \quad x \in [0, 5],
\end{align} 
subject to the initial condition $s(x,0) = u(x)$.
For single solitons, the exact solution can be derived as \cite{brauer2000korteweg}
\begin{align}
    \label{eq: KDV_exact_sol}
    s(x, t) = \frac{c}{2}  \operatorname{sech}^{2}\left[\frac{\sqrt{c}}{2}(5 x - \frac{1}{10} c t - a)\right]
\end{align}
where $c$ denotes the travel speed of a wave and $a$ determines the initial wave position in the physical domain. It is well-known that the KDV equation is one of the most important non-linear PDEs, originally derived to model shallow water waves and then used to describe a diversity of important finite amplitude dispersive wave phenomena in physics, such as acoustic waves in a harmonic crystal and ion-acoustic waves in plasmas \cite{miles1981korteweg}. Using Algorithm \ref{alg: long_time_integration},
the problem can be reduced to learning the solution operator $G$ mapping initial conditions to the corresponding PDE solutions in $[0, \Delta t]$.

We proceed by approximating $G$ with a DeepONet $G_{\bm{\theta}}$, where both the branch and trunk networks are 7-layer modified fully-connected neural networks (see euqations (\ref{eq:ADGM_1})-(\ref{eq:ADGM_2})) with 200 neurons per hidden layer. This allows us to define the corresponding PDE residual
\begin{align}
    \mathcal{R}[\bm{u}](x, t) = \frac{\partial G_{\bm{\theta}  } (\bm{u})}{\partial t}(x, t) + \epsilon  G_{\bm{\theta}}  (\bm{u})(x, t) \frac{\partial G_{\bm{\theta}  } (\bm{u})}{\partial x}(x, t) + \mu \frac{\partial^3 G_{\bm{\theta}}  (\bm{u})}{\partial x^3}(x, t),  
\end{align}
used to formulate the following loss function
\begin{align}
    \mathcal{L}(\bm{\theta}) =  \mathcal{L}_{\text{data}}(\bm{\theta}) +  \mathcal{L}_{\text{physics}}(\bm{\theta}),
\end{align}
where 
\begin{align*}
    &\mathcal{L}_{\text{data}}(\bm{\theta}) = \frac{1}{NP} \sum_{i=1}^N \sum_{j=1}^P \left| G_{\bm{\theta}}(\bm{u}^{(i)})(x_{s,j}^{(i)}, t_{s, j}^{(i)}) - s^{(i)}(x_{s,j}^{(i)}, t_{s, j}^{(i)})  \right|, \\
    &\mathcal{L}_{\text{physics}}(\bm{\theta})  =  \frac{1}{NQ} \sum_{i=1}^N \sum_{j=1}^Q \left| \mathcal{R}[\bm{u}^{(i)}](x_j^{(i)}, t_j^{(i)}) - s^{(i)}(x_{r,j}^{(i)}, t_{r,j}^{(i)})   \right|.
\end{align*}
Here, $\bm{u}^{(i)} = [u^{(i)}(x_1), u^{(i)}(x_2), \dots, u^{(i)}(x_m)]$ denotes the initial conditions evaluated at a set of fixed sensor locations $\{x_i\}_{i=1}^m$, and $s^{(i)}$ is the associated PDE solution corresponding to each $u^{(i)}$. For every $i$, $\{(x_{s,j}^{(i)}, t_{s, j}^{(i)}), s^{(i)}(x_{s,j}^{(i)}, t_{s, j}^{(i)})\}_{j=1}^P$ are available solution measurements, while $\{(x_j^{(i)}, t_j^{(i)}) - s^{(i)}(x_{r,j}^{(i)}, t_{r,j}^{(i)}) \}_{j=1}^Q$ is a set of collocation points randomly sampled in the computational domain $[0,5] \times [0, \Delta t]$ for imposing the PDE constraint. To obtain a set of training data, we  sample  pairs $(a, c)$ where $a \sim \mathcal{U}(0, 5), b \sim \mathcal{U}(0, 1)$, and randomly selecting $P$ observations of the exact solution $s(x,t)$ using equation (\ref{eq: KDV_exact_sol}). In this example, we set $N = 5 \times 10^3$, $\Delta t = 10$ and $m = P = Q = 200$. 

We train the proposed physics-informed DeepONet using the Adam optimizer for $2 \times 10^5$ iterations, and then apply Algorithm \ref{alg: long_time_integration} to the initial condition with $a=1, c = 3/2$, i.e 
\begin{align*}
    s(x,0) =  \frac{3}{4}  \operatorname{sech}^{2}\left[\frac{\sqrt{3/2}}{2}(5 x - 1 )\right].
\end{align*}

\begin{figure}
     \centering
     \begin{subfigure}[b]{0.7\textwidth}
         \centering
    \includegraphics[width=\textwidth]{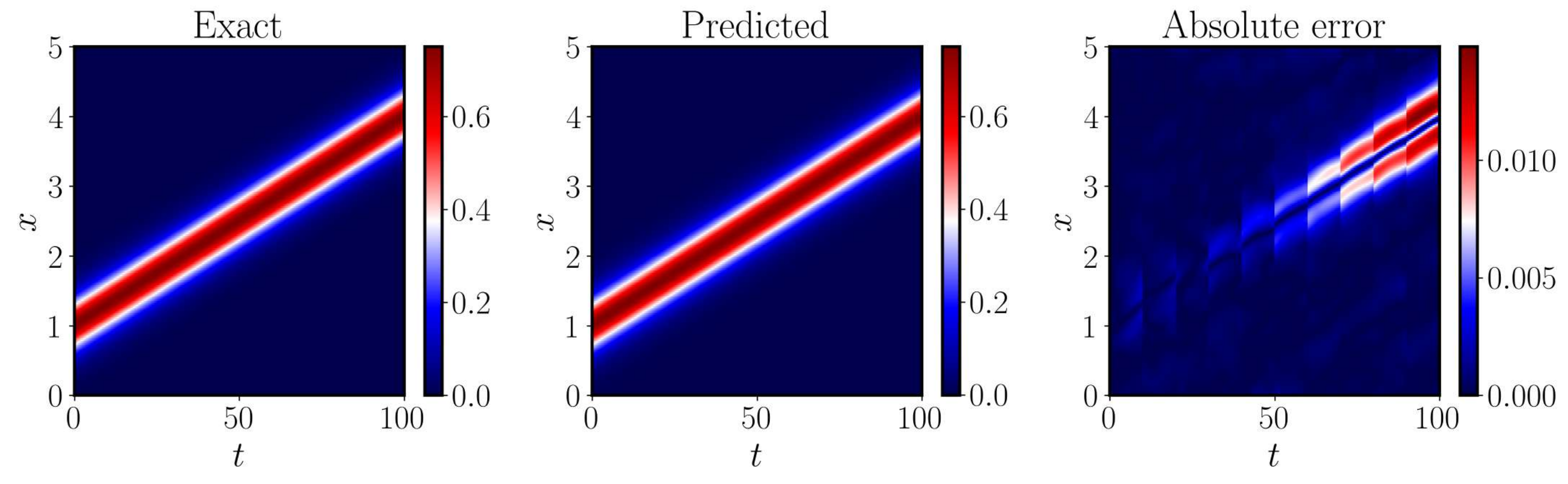}
     \end{subfigure}
     \begin{subfigure}[b]{0.7\textwidth}
         \centering
    \includegraphics[width=\textwidth]{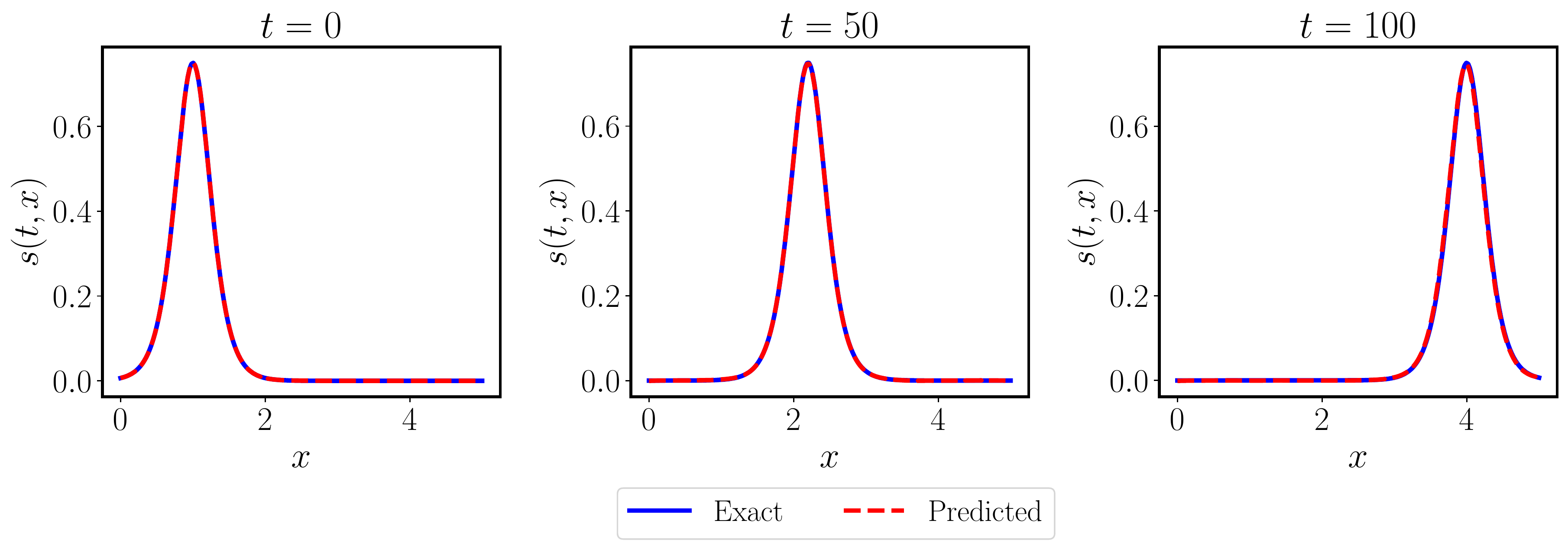}
     \end{subfigure}
        \caption{{\em Korteweg–De Vries equation:} {\em Top:} Exact solution versus the predicted solution of a trained physics-informed 
        DeepONet using Algorithm \ref{alg: long_time_integration} for integrating the PDE system (\ref{eq: KDV}) up to $T = 100$. The relative $L^2$ error is $1.24\%$. {\em Bottom:} Comparison between the exact and the predicted solutions at different temporal snapshots corresponding to $t=0, 50, 100$. }
        \label{fig: PI_deeponet_KDV}
\end{figure}

\begin{figure}
     \centering
     \begin{subfigure}[b]{0.7\textwidth}
         \centering
    \includegraphics[width=\textwidth]{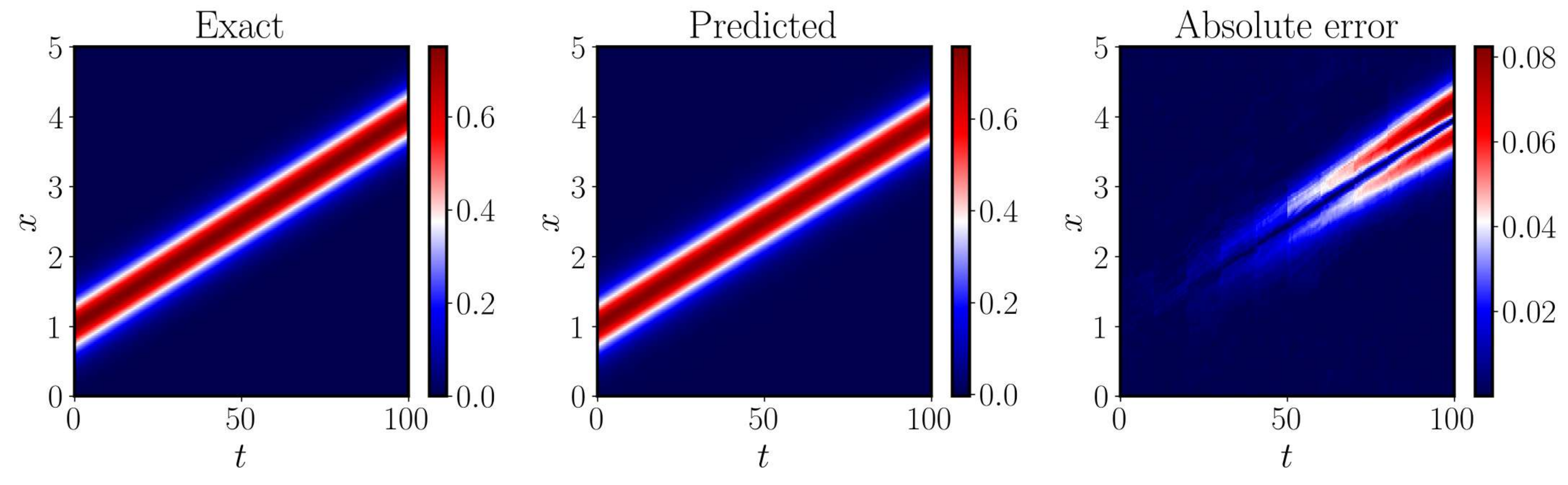}
     \end{subfigure}
     \begin{subfigure}[b]{0.7\textwidth}
         \centering
    \includegraphics[width=\textwidth]{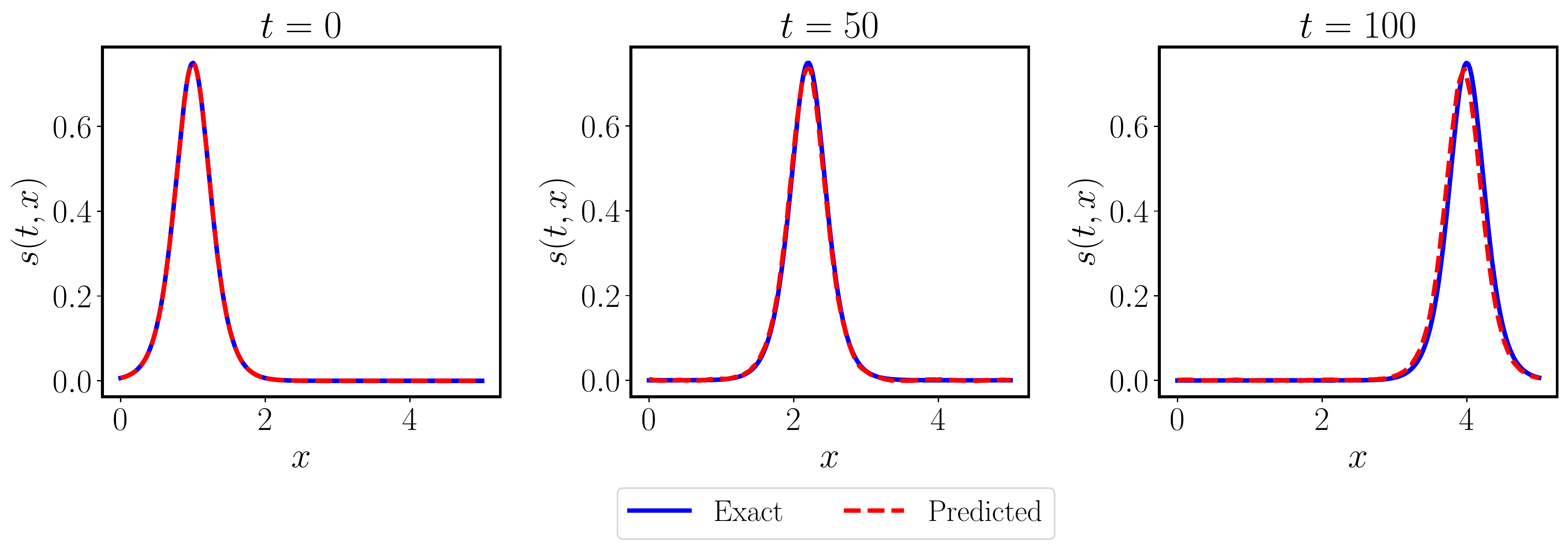}
     \end{subfigure}
   \caption{{\em Korteweg–De Vries equation:} {\em Top:} Exact solution versus the predicted solution of a trained conventional DeepONet using Algorithm \ref{alg: long_time_integration} for integrating the PDE system (\ref{eq: KDV}) up to $T = 100$. The relative $L^2$ error is $6.84\%$. {\em Bottom:} Comparison between the exact and the predicted solutions at different temporal snapshots corresponding to $t=0, 50, 100$. }
        \label{fig: deeponet_KDV}
\end{figure}

\begin{figure} 
    \centering
    \includegraphics[width=0.4\textwidth]{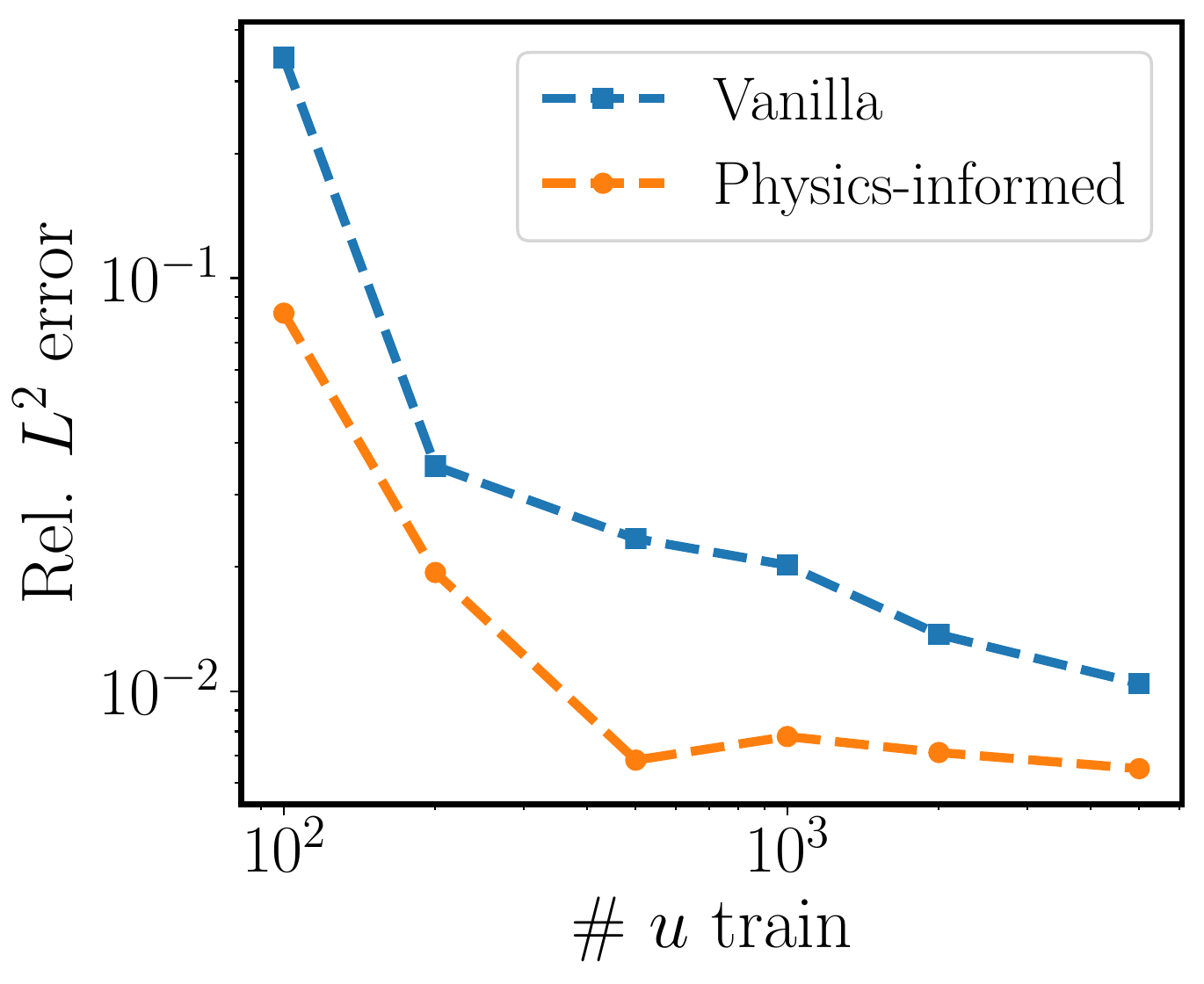}
    \caption{Relative $L^2$ prediction error of a vanilla DeepONet \cite{lu2021learning} versus the proposed physics-informed DeepONet as a function of the number of examples in the training data-set.}
    \label{fig: KDV_error}
\end{figure}

The top panel of Figure \ref{fig: PI_deeponet_KDV} shows the predicted global  spatio-temporal solution, for which the resulting prediction error is measured at $1.24e-02$ in the relative $L^2$-norm.  A more detailed assessment of the predicted solution is presented in the bottom panel of  Figure \ref{fig: PI_deeponet_KDV}, which displays  a comparison between the exact and the predicted solutions at different temporal snapshots $t=0, 50, 100$. Through  domain decomposition in time, the physics-informed DeepONet can accurately predict the traveling wave across a long-time horizon. We further investigate the performance of a conventional DeepONet model, which can be trained by minimizing the loss function $\mathcal{L}_{\text{data}}$ solely.
As shown in Figure \ref{fig: deeponet_KDV}, the results of the vanilla DeepONet are slightly worse than the physics-informed DeepONet. This conclusion is further confirmed by a relatively large prediction error of $6.84e-02$. 
For a more thorough comparison, we train both a vanilla DeepONet and a physics-informed DeepONet for different number of training data points
(i.e, different number of samples $u$), and report the mean of the relative $L^2$  error of $s(x,t)$ over all 100 examples in the
test data-set in Figure \ref{fig: KDV_error}. Compared to vanilla DeepONets, the proposed physics-informed DeepONets can not only achieve better predictive accuracy, but also utilize the observed data more efficiently, therefore providing enhanced effectiveness in small data regime.


\section{Discussion}\label{sec:discussion}


We have presented an effective methodology for performing long-time integration of evolution equations, for which many popular ML-based approaches such as physics-informed neural networks (PINNs) often struggle to yield accurate results. The proposed approach avoids the high computational cost associated with training multiple networks \cite{meng2020ppinn, wight2020solving,du2021evolutional,raissi2019deep}, and introduces a new effective way to temporal domain decomposition in which a single network needs to be trained only within a short-time interval, albeit across a distribution of initial conditions. Leveraging the framework of physics-informed DeepOnets \cite{wang2021learning}, we put forth a simple two-step process. First, we demonstrate how deep neural networks can parametrize and learn solution operators that map initial conditions to associated ODE/PDE solutions in a short-time interval. The trained can be then iteratively evaluated to construct the global ODE/PDE in large temporal domains, at any arbitrary spatio-temporal resolution, and across a range of initial conditions.
The effectiveness and robustness of the proposed algorithms has been demonstrated in a series of detailed numerical experiments involving long-time simulation of evolution laws that describe inherently different physical processes, including wave propagation, reaction-diffusion dynamics, and stiff chemical kinetics.

Long-time integration is often one of the main bottlenecks in simulating complex multi-scale and multi-physics processes in science and engineering, with applications ranging from understanding the effect of anthropogenic pollution in ocean and atmospheric chemical transport, to designing fuel-efficient combustion engines, to elucidating the biophysical mechanisms underpinning cardiovascular disease, and beyond. Developing high-fidelity simulation tools for such problems is technically and computationally challenging due to the long-time duration of interest, the stochastic nature of fragmentation and turbulent mixing phenomena, the temperature dependency of thermal neutralization mechanisms, and the stiff dynamics of bio-chemical reactions. The methods presented in this work are a first step towards demonstrating feasibility for ML-based techniques in reducing computational costs and enabling the rapid and accurate emulation of such  non-equilibrium processes in science and engineering.

Despite the early promise demonstrated here, we have to admit that we are still at the very early stages of tackling long-time prediction problems with physics-informed DeepONets. There are  many open questions worth considering as future research directions. From a theoretical point of view, it is important to develop a better understanding of how approximation errors affect the stability  and accuracy of the proposed methods. This is a crucial element in applications that demand accuracy and convergence guarantees, where currently classical numerical solvers remain the de-facto choice. From a practical point of view, it would be intriguing to apply the proposed approach to solve chaotic dynamical systems, such as  Kuramoto-Sivashinsky equation \cite{sivashinsky1977nonlinear, kuramoto1978diffusion} or the Navier-Stokes equations in the turbulent regime. These problems typically involve fast transitions of frequencies and are extremely sensitive to initial conditions, which inevitably introduces great challenges to both traditional numerical solvers, as well as ML-based approaches. From a methodology point of view, our algorithm can be regarded as a special "domain decomposition" in time, which further motivates us to develop domain decomposition in space with physics-informed DeepONets. This may be combined with  traditional domain decomposition techniques such as the Schwartz alternating method \cite{lions1988schwarz} to open the path of scaling physics-informed ML approaches to large computational domains and complex geometries \cite{wang2021train}. We believe that addressing these open questions will pave a new way to developing scientific machine learning algorithms with better robustness and accuracy guarantees, as needed for many critical applications in computational science and engineering.

\section*{Author Contributions}
SW and PP conceptualized the research and designed the numerical studies. SW  implemented the methods and conducted the numerical experiments. PP provided funding and supervised all aspects of this work. All authors contributed in writing the manuscript.

\section*{Acknowledgements}
This work received support from DOE grant DE-SC0019116, AFOSR grant FA9550-20-1-0060, and DOE-ARPA grant DE-AR0001201. We would also like to thank the developers of the software that enabled our research, including JAX \cite{jax2018github}, Matplotlib \cite{hunter2007matplotlib},  NumPy \cite{harris2020array} and DeepXDE \cite{lu2021deepxde}.

\bibliographystyle{unsrt}
\bibliography{references}

\clearpage

\appendix

\input{Appendix}

\end{document}

%% file: Appendix.tex
\section{Notations}
Table \ref{tab: Notations} summarizes the main symbols and notation used in this work.

\begin{table}[h]
\renewcommand{\arraystretch}{1.4}
    \centering
    \begin{tabular}{ll} 
    \Xhline{3\arrayrulewidth} 
    Notation     & Description \\
    \Xhline{3\arrayrulewidth} 
        $\bm{u}(\cdot)$ & an input function \\  
        $\bm{s}(\cdot)$ & a solution to a parametric PDE \\
        $G$         & an operator \\
        $G_{\bm{\theta}}$  &  an DeepONet representation of the operator $G$ \\
        $\bm{\theta}$ &  all trainable parameters of a DeepONet \\
        $\{\bm{x}_i\}_{i=1}^m$  & $m$ sensor points where input functions $\bm{u}(\bm{x})$ are evaluated\\
        $[u(\bm{x}_1), u(\bm{x}_2), \dots, u(\bm{x}_m)]$ & an input of the branch net, representing the input function $u$  \\
        N                & \# input samples in the training data-set \\
        m               &  \# locations for evaluating the input functions $u$        \\
        P               &   \# locations for evaluating the output functions $G(u)$        \\
        Q                &  \# collocation points for evaluating the PDE residual \\
        GRF             &  a Gaussian random field                   \\
        $l$              & length scale of a Gaussian random field   \\
    \Xhline{3\arrayrulewidth}
    \end{tabular}
    \caption{{\em Nomenclature}: Summary of the main symbols and notation used in this work.}
    \label{tab: Notations}
\end{table}

\section{Hyper-parameter settings}

\label{sec: parameters}
Table \ref{tab: parameters_case} summarizes the hyper-parameter setting for all examples considered in this work.

\begin{table}[h]
\renewcommand{\arraystretch}{1.4}
    \centering
    \begin{tabular}{c|ccccccc}
        \Xhline{3\arrayrulewidth}
      Case   & Input function space &  m & P & Q & \#u Train  & \# u Test & Iterations  \\
      \hline
     Gravity pendulum & $\mathcal{U}(-3, 3)$ & 1 & 1 & 100 & $5 \times 10^4$ & 100  & $3 \times 10^{5}$ \\
    Linear ODE &  GRF ($l = 0.5$) \& $\mathcal{U}(-2, 2)$ & 100 \& 1 & 1 & 100 & $5 \times 10^4$ & 100  & $2 \times 10^{5}$ \\
    Stiff ODE & $\mathcal{U}(0, 1)$ & 1 & 1 & $10^3$ & $5 \times 10^4$ & 100  & $4 \times 10^{5}$ \\
    Wave equation & GRF($l=0.5$) & 100 & 100 & 200 & $ 10^4$ & 100  & $2 \times 10^{5}$ \\
    Diffusion-reaction equation & $GRF($l=0.2$)$ & 100 & 100 & $100$ & $ 10^4$ & 100  & $2 \times 10^{5}$ \\
    KDV equation & $\mathcal{U}(1, 2) \times \mathcal{U}(0, 5)$ & 200 & 200 & 200 & $5 \times 10^3$ & 100  & $2 \times 10^{5}$ \\
    \Xhline{3\arrayrulewidth}
    \end{tabular}
    \caption{Default hyper-parameter settings for each benchmark employed in this work  (unless otherwise stated).}
    \label{tab: parameters_case}
\end{table}

\begin{table}[h]
\renewcommand{\arraystretch}{1.4}
    \centering
    \begin{tabular}{c|cccccc}
        \Xhline{3\arrayrulewidth}
        Case   &   Trunk depth &  Trunk width & Branch depth & Branch width  \\
        \hline
        Gravity pendulum & 8 & 100  &   8 & 100  \\
        Linear ODE & 7 & 100  &   7 & 100  \\
         Stiff ODE & 7 & 100  &   7 & 100  \\
        Wave equation & 5 & 200  &   5 & 200  \\
        Diffusion-reaction equation  & 5 & 100  &  5 & 100  \\
       KDV equation  & 7 & 200  &  7 & 200  \\
         \Xhline{3\arrayrulewidth}
    \end{tabular}
    \caption{DeepONet architectures for each benchmark employed in this work (unless otherwise stated). }
    \label{tab: Physics_informed_DeepONet_size}
\end{table}

\section{Performance metrics}
\label{sec: test_error}

The error metric employed throughout all numerical experiments to assess model performance is the relative $L^2$ norm. Specifically, the reported test errors correspond to the mean of the relative $L^2$ error of a trained physics-informed DeepONet model over all examples in the test data-set, i.e
\begin{align}\label{eq: test_error}
    \text{Test error} := \frac{1}{N} \sum_{i=1}^N \frac{\|G_{\theta}(\bm{u}^{(i)})(y) - G(\bm{u}^{(i)})(y)\|_2 }{\|G(\bm{u}^{(i)})(y)\|_2},
\end{align}
where $N$ denotes the number of examples in the test data-set and $y$ is typically a set of equi-spaced points in the domain of $G(u)$. Here $G_{\theta}(\bm{u}^{(i)})(y)$ denotes the predicted DeepONet outputs, while $G(\bm{u}^{(i)})(y)$ corresponds to the ground truth target functions.

\section{Computational cost}

{\bf Training:} Table \ref{tab: computational_cost} summarizes the computational cost  (hours) of training different models
The size of different models as well as network architectures are listed Table  \ref{tab: Physics_informed_DeepONet_size}. All networks are trained using a single A100 card. It can be observed that training a physics-informed DeepONet
model is generally slower than  training a conventional DeepONet. This is expected as physics-informed DeepONets require to compute the PDE residual via automatic differentiation, yielding a lager computational graph, and, therefore, a higher computational cost.

\begin{table}[h]
\renewcommand{\arraystretch}{1.4}
    \centering
    \begin{tabular}{c|c| c}
     \Xhline{3\arrayrulewidth}
       Case  &  Model &  Training time (hours)  \\
       \hline
      \multirow{2}{*}{ Gravity pendulum} & Physics-informed neural network  & 0.12
      \\ &Physics-informed DeepONet  &  1.63   \\
      \hline
        Linear ODE & Physics-informed DeepONet  &  1.33 \\
        \hline
    Stiff ODE & Physics-informed DeepONet  &  7.60 \\
        \hline
    Wave equation & Physics-informed DeepONet  &  3.00 \\
    Diffusion-reaction equation & Physics-informed DeepONet  & 1.48  \\
        \hline
    \multirow{2}{*}{ KDV equation} & Physics-informed DeepONet  & 2.17
      \\ &DeepONet  & 0.35    \\
    \Xhline{3\arrayrulewidth}
    \end{tabular}
    \caption{Computational cost (hours) for training different models across the different benchmarks and architectures employed in this work. Reported timings are obtained on a single Nvidia V100 GPU.}
    \label{tab: computational_cost}
\end{table}

{\bf Inference:} Once trained, the model can be rapidly queried to return prediction at short-time intervals. Each query typically takes $\mathcal{O}(10^{-3}\text{ms})$  on a single Nvidia V100 GPU. Constructing the global PDE/ODE solution in large temporal domains using Algorithm \ref{alg: long_time_integration} requires $N\sim\mathcal{O}(10)- \mathcal{O}(10^3)$ evaluations of the trained model, typically leading to a total inference time of $\sim \mathcal{O}(10^{-2})$ sec for a given initial condition. Multiple initial conditions can also be simulated at once, as our JAX \cite{jax2018github} implementation is trivial vectorized and parallelized on GPU hardware. As demonstrated in Figure \ref{fig: PI_deeponet_stiff_ODE_s_error_time}, the proposed framework can perform long-time integration of more than $1,000$ initial conditions in $\mathcal{O}(1)$ second, yielding a $\sim$10x-50x speedup compared to a traditional numerical solver. Unlike traditional numerical solvers that are heavily specialized to a specific type of dynamic behavior, this cost remains nearly constant for all examples considered in this work, regardless of the ODE/PDE system that is simulated, as it merely amounts to the cost associated with evaluating the forward pass of the trained DeepONet model.

\clearpage
\section{Supplementary Figures}

\subsection{Gravity pendulum}



\begin{figure}[h]
     \centering
     \begin{subfigure}[b]{0.4\textwidth}
         \centering
    \includegraphics[width=1.0\textwidth]{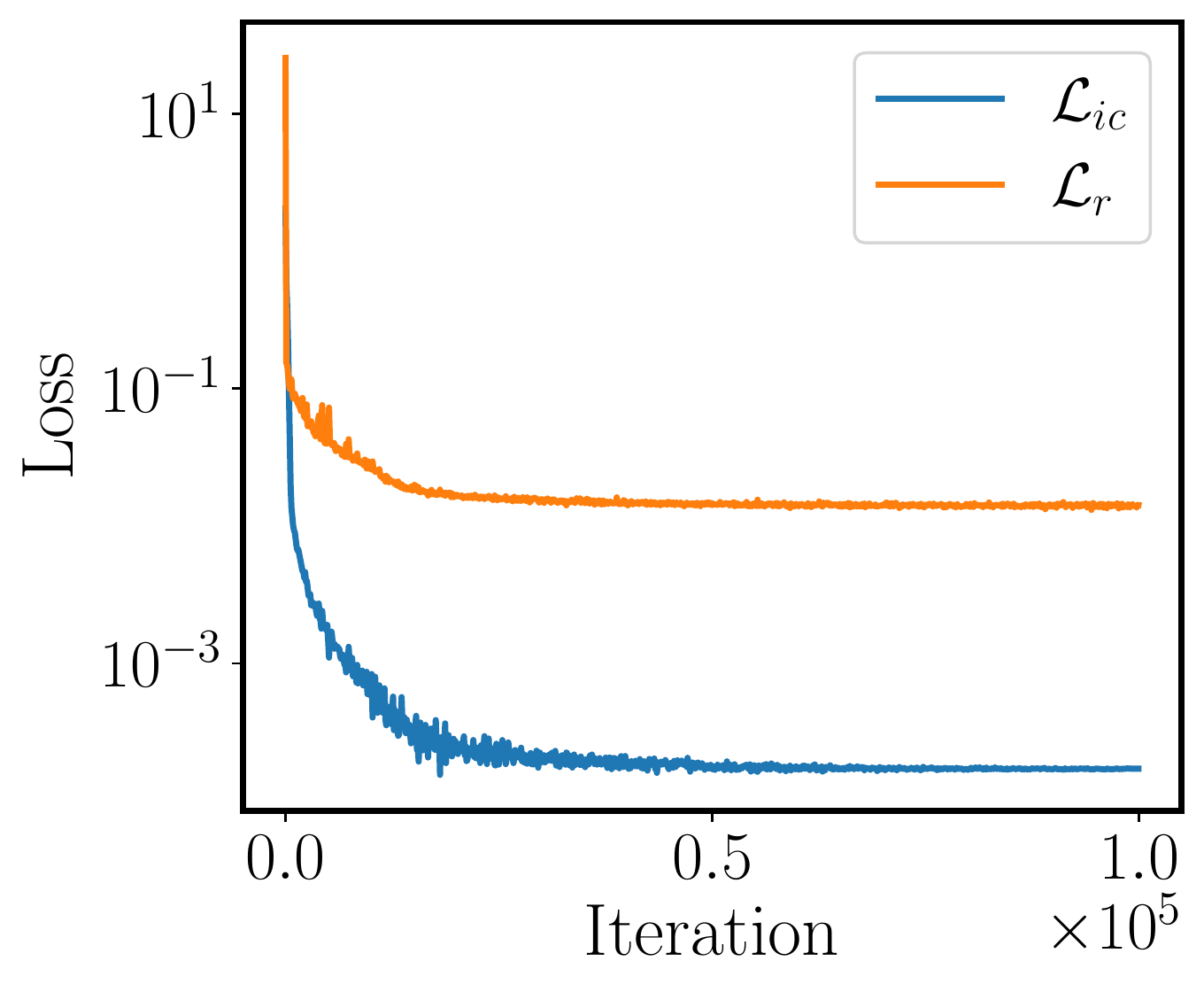}
    \label{fig: PINN_pendulum_loss}
     \end{subfigure}
        \begin{subfigure}[b]{0.4\textwidth}
          \centering
    \includegraphics[width=1.0\textwidth]{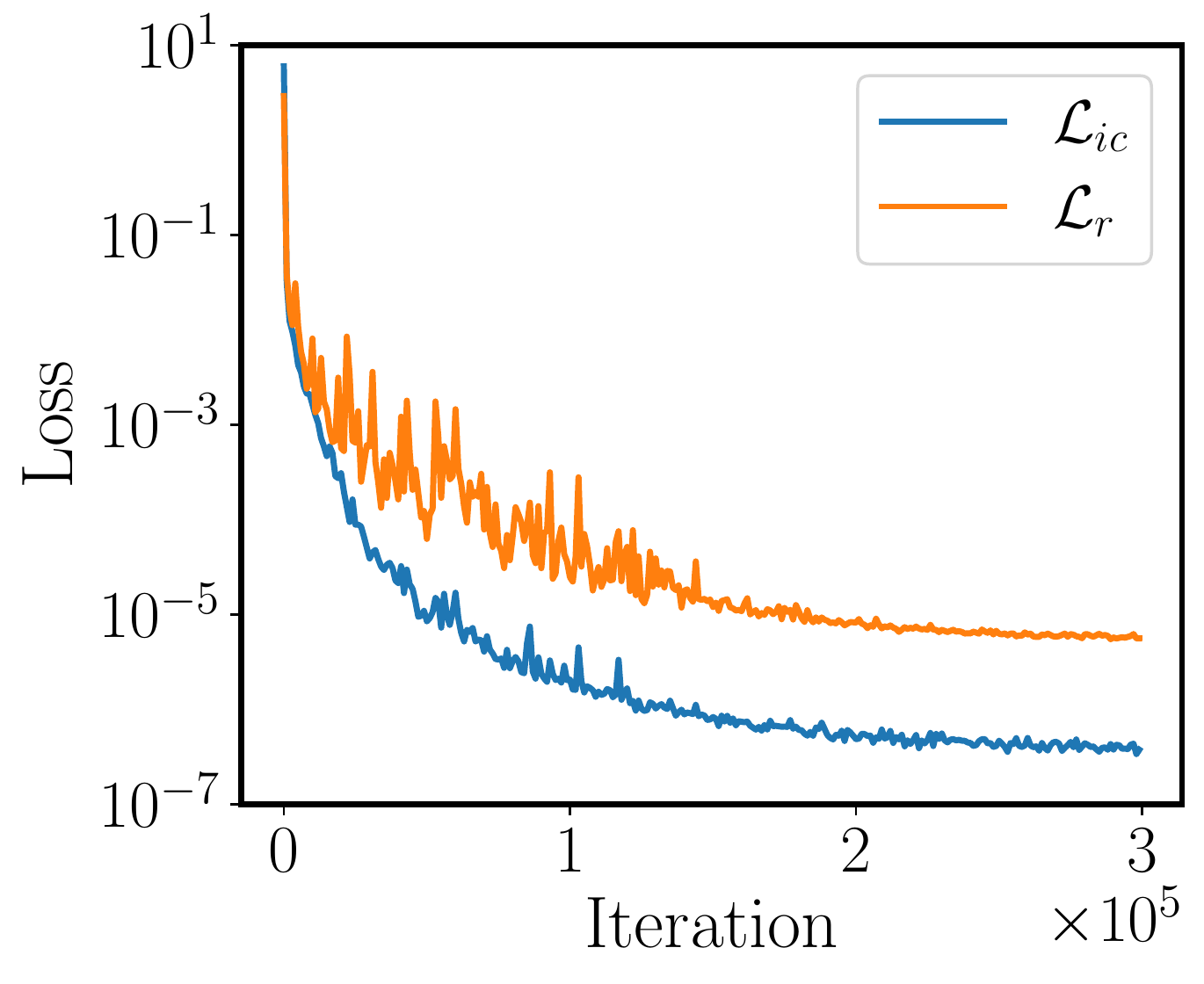}
    \caption{}
    \label{fig: PI_deeponet_pendulum_loss}
     \end{subfigure}
       \caption{{\em Gravity pendulum:} {\em Left:} Training loss convergence of a conventional PINN modelfor $10^5$ iterations of gradient descent using the Adam optimizer. {\em Right:}  Training loss convergence of a  physics-informed DeepONet model for $3 \times 10^5$ iterations of gradient descent using the Adam optimizer.}
        \label{fig: pendulum_losses}
\end{figure}

\clearpage
\subsection{Inhomogeneous ODE}

\begin{figure}[h]
    \centering
    \includegraphics[width=0.4\textwidth]{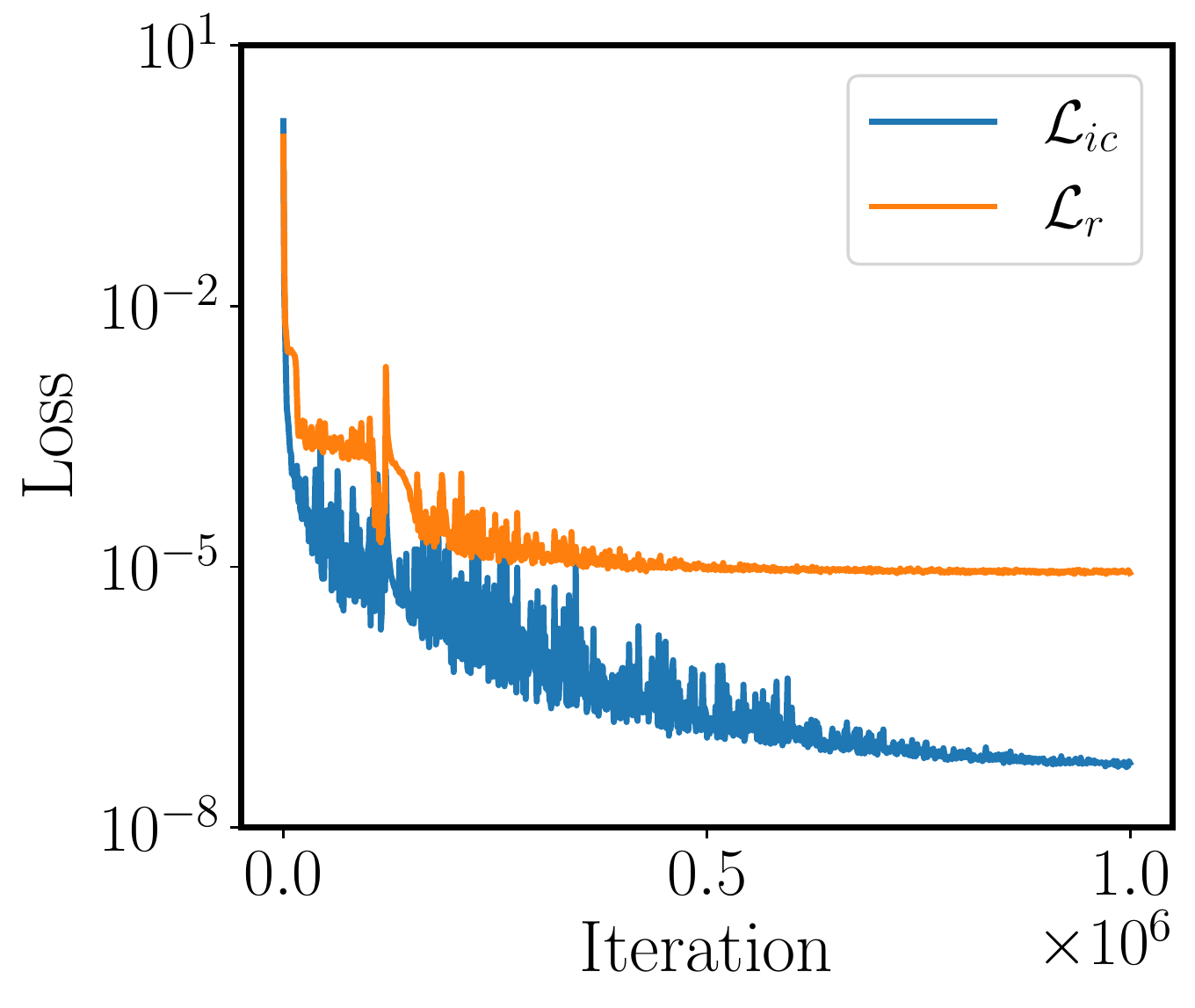}
    \caption{{\em Linear ODE:}  Training loss convergence of a  physics-informed DeepONet model for $10^5$ iterations of gradient descent using the Adam optimizer.}
    \label{fig: PI_deeponet_ODE_loss}
\end{figure}

\begin{figure}[h]
     \centering
     \begin{subfigure}[b]{0.8\textwidth}
         \centering
         \includegraphics[width=\textwidth]{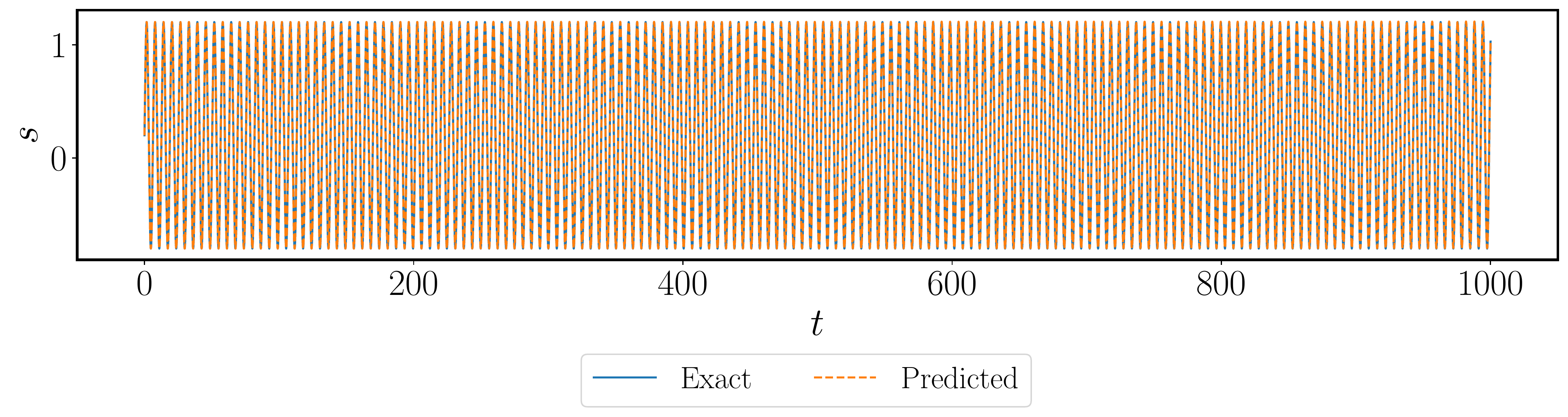}
     \end{subfigure}
          \begin{subfigure}[b]{0.8\textwidth}
         \centering
         \includegraphics[width=\textwidth]{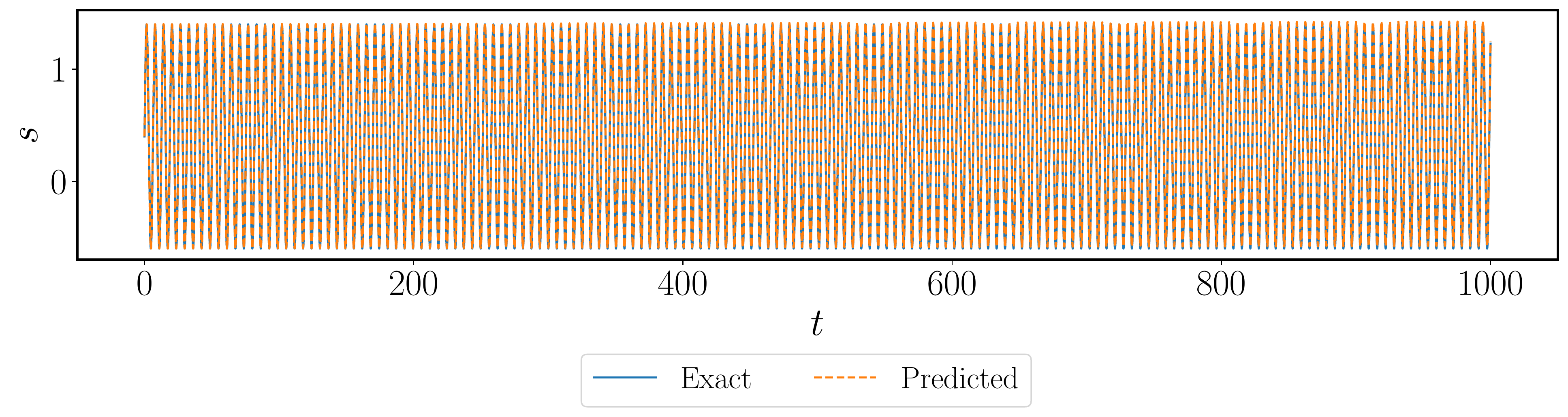}
     \end{subfigure}
               \begin{subfigure}[b]{0.8\textwidth}
         \centering
         \includegraphics[width=\textwidth]{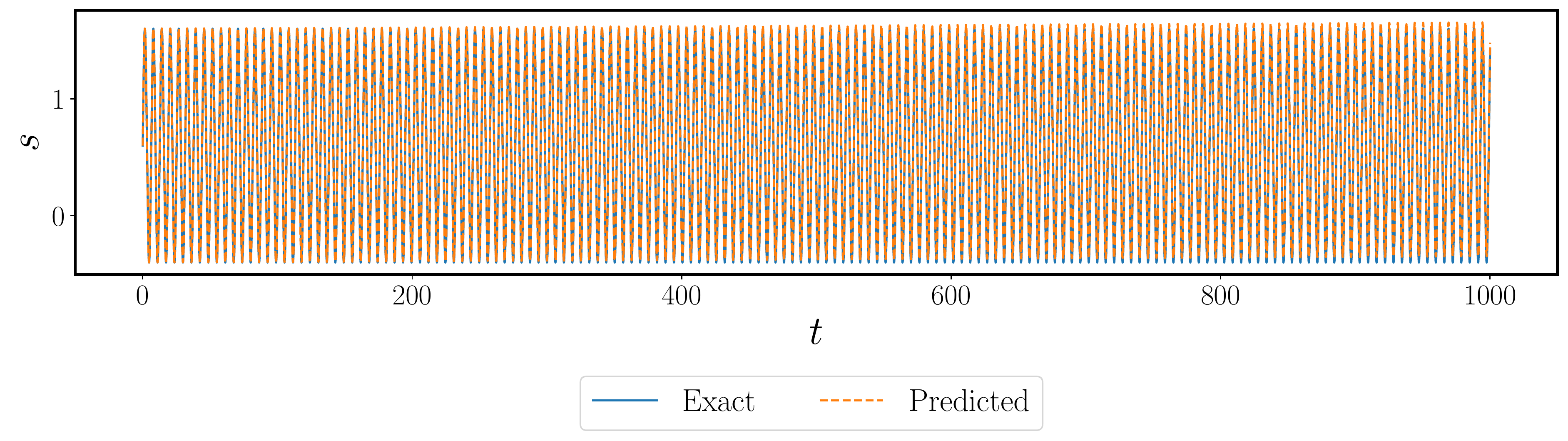}
     \end{subfigure}
        \caption{{\em Linear ODE:} Predicted solutions obtained by  a trained physics-informed DeepONet using Algorithm \ref{alg: long_time_integration}, across three different initial conditions $s(0) = 0.2, 0.4, 0.6$. The relative $L^2$ errors are $0.52\%, 1.79\%, 3.30\%$ respectively.  }
        \label{fig: PI_deeponet_ODE_s_pred_examples}
\end{figure}

\clearpage
\subsection{Stiff chemical kinetics}


\begin{figure}[h]
     \centering
     \begin{subfigure}[b]{0.35\textwidth}
         \centering
    \includegraphics[width=\textwidth]{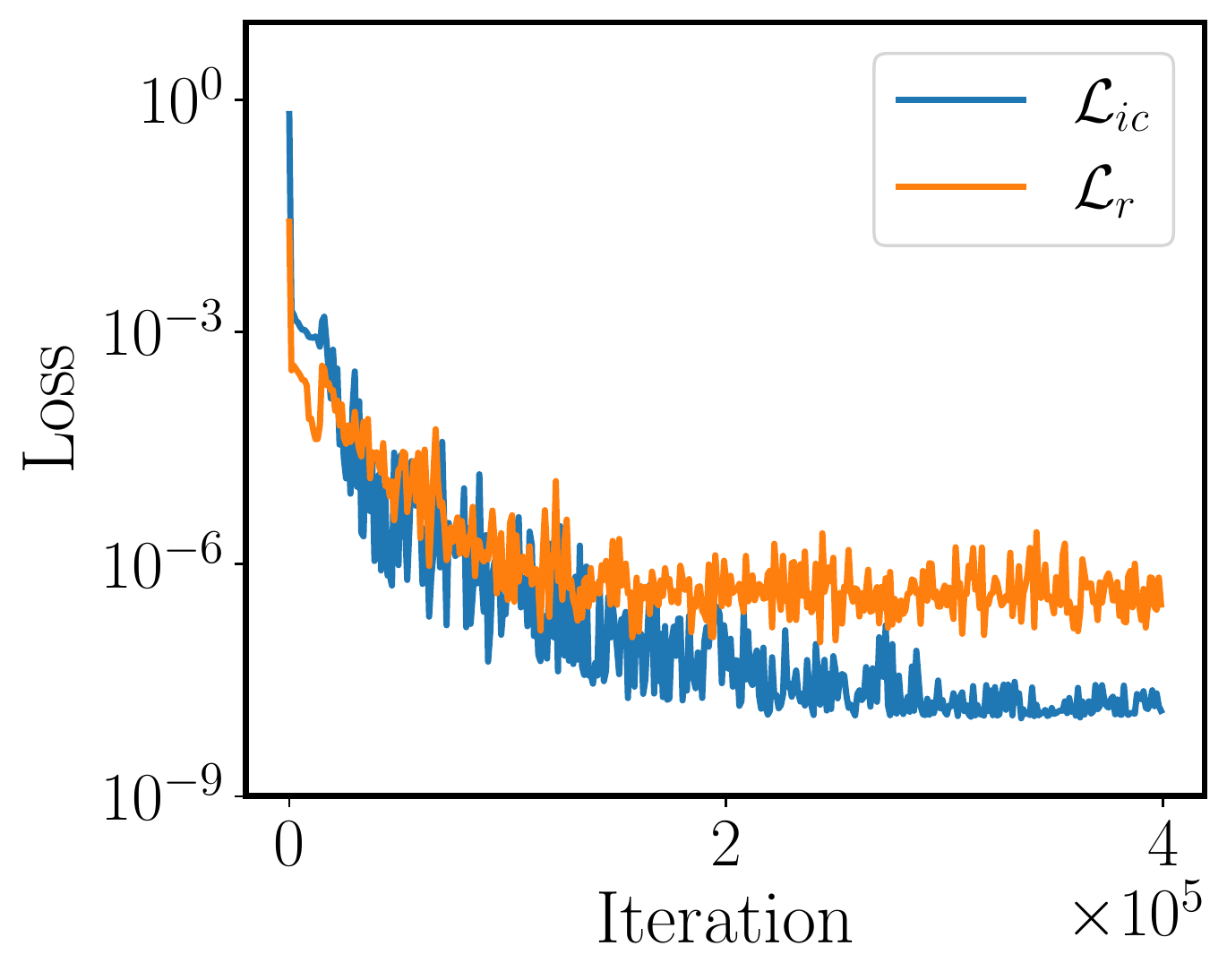}
         \label{fig: PI_deeponet_stiff_ODE_loss}
     \end{subfigure}
     \begin{subfigure}[b]{0.35\textwidth}
         \centering
    \includegraphics[width=\textwidth]{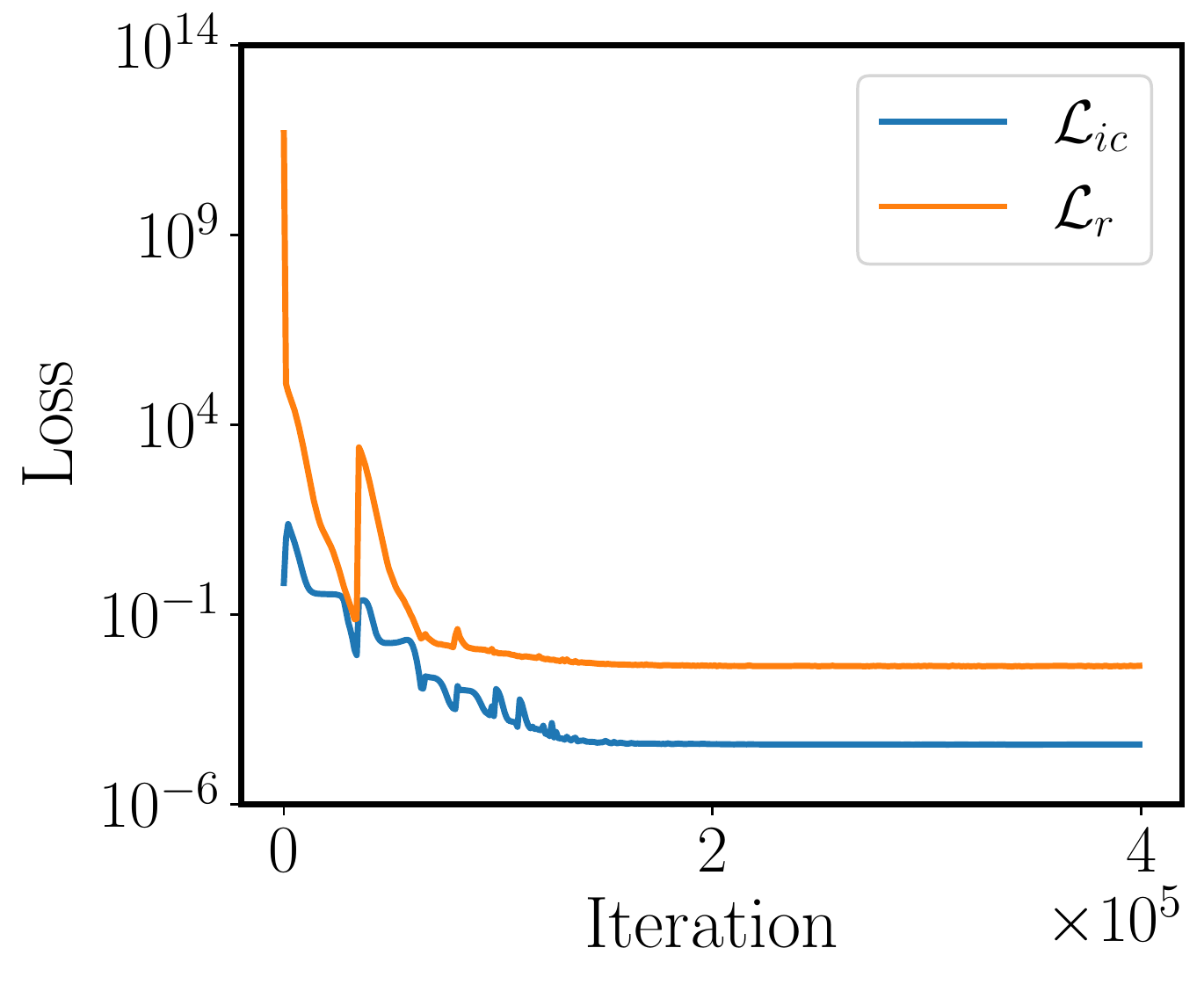}
         \label{fig: PI_deeponet_unscaled_stiff_ODE_loss}
     \end{subfigure}
        \caption{{\em Stiff ODE:} {\em Left:} Training loss convergence of a  scaled physics-informed DeepONet model for $4 \times 10^5$ iterations of gradient descent using the Adam optimizer. {\em Right:}  Training loss convergence of a unscaled physics-informed DeepONet model for $4 \times 10^5$ iterations of gradient descent using the Adam optimizer.}
        \label{fig: PI_deeponet_stiff_ODE_losses}
\end{figure}

\begin{figure}[h]
     \centering
     \begin{subfigure}[b]{0.6\textwidth}
         \centering
         \includegraphics[width=\textwidth]{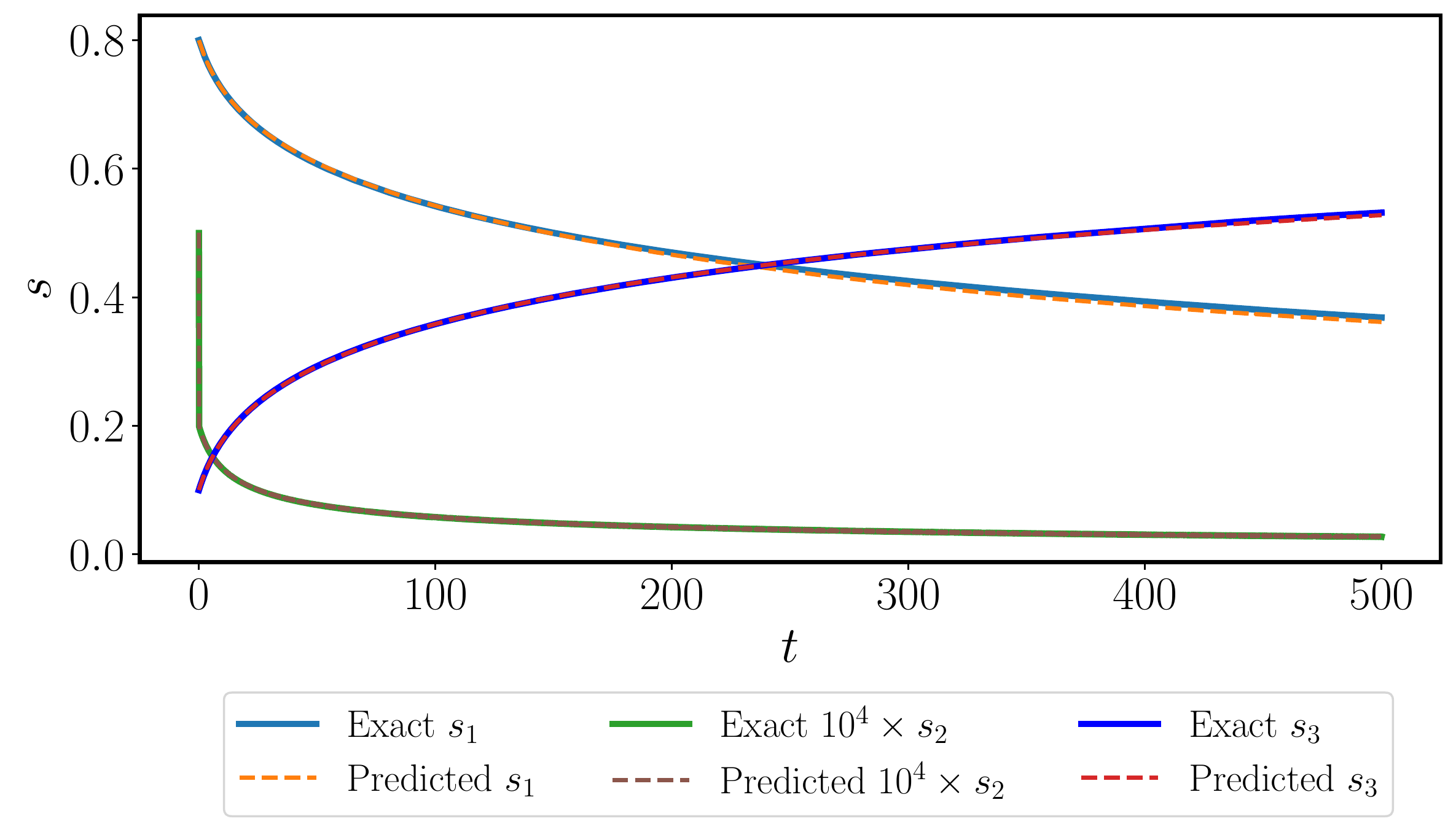}
         \label{fig: PI_deeponet_stiff_ODE_s_pred_1}
     \end{subfigure}
          \begin{subfigure}[b]{0.6\textwidth}
         \centering
         \includegraphics[width=\textwidth]{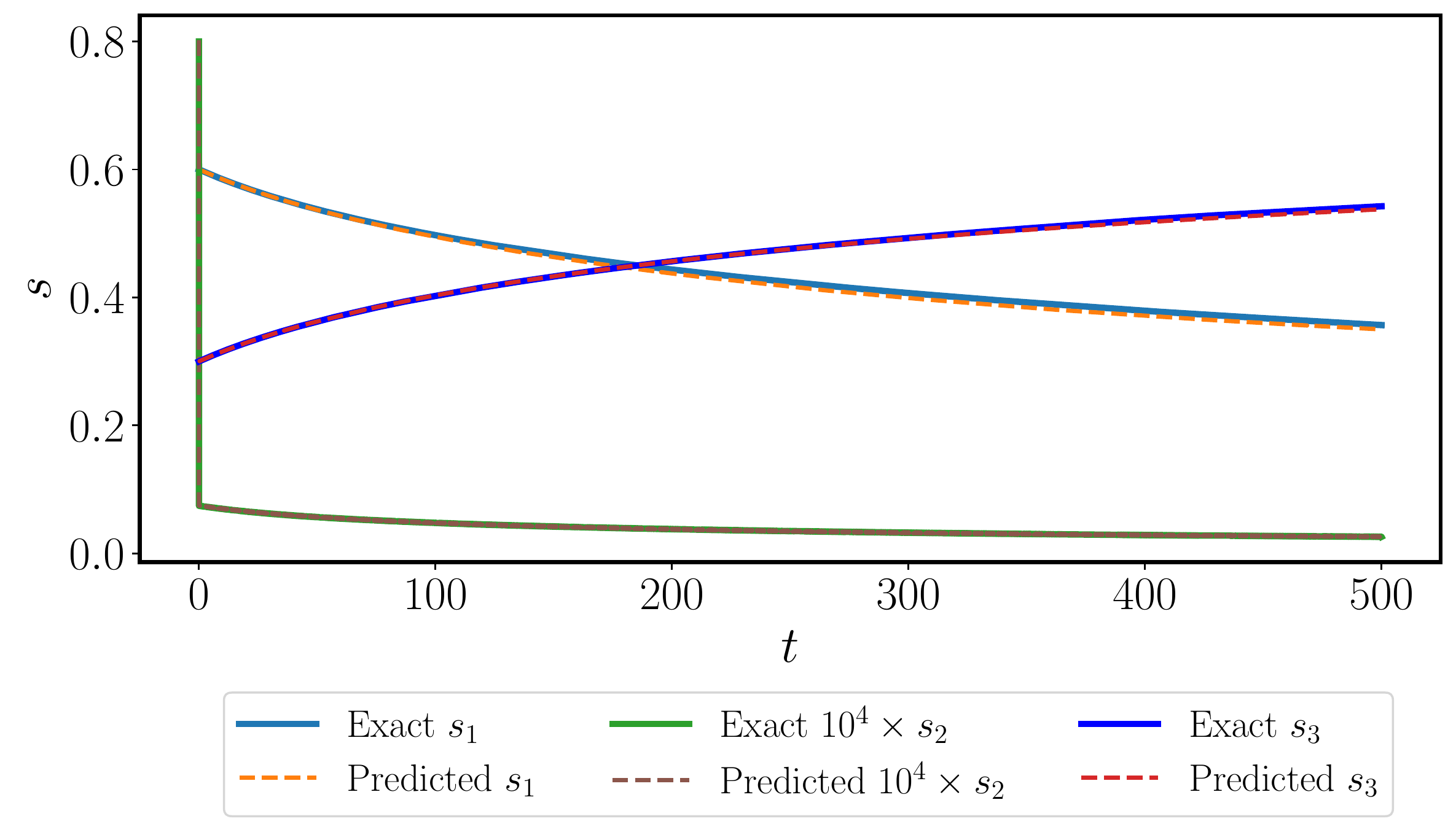}
         \label{fig: PI_deeponet_stiff_ODE_s_pred_2}
     \end{subfigure}
               \begin{subfigure}[b]{0.6\textwidth}
         \centering
         \includegraphics[width=\textwidth]{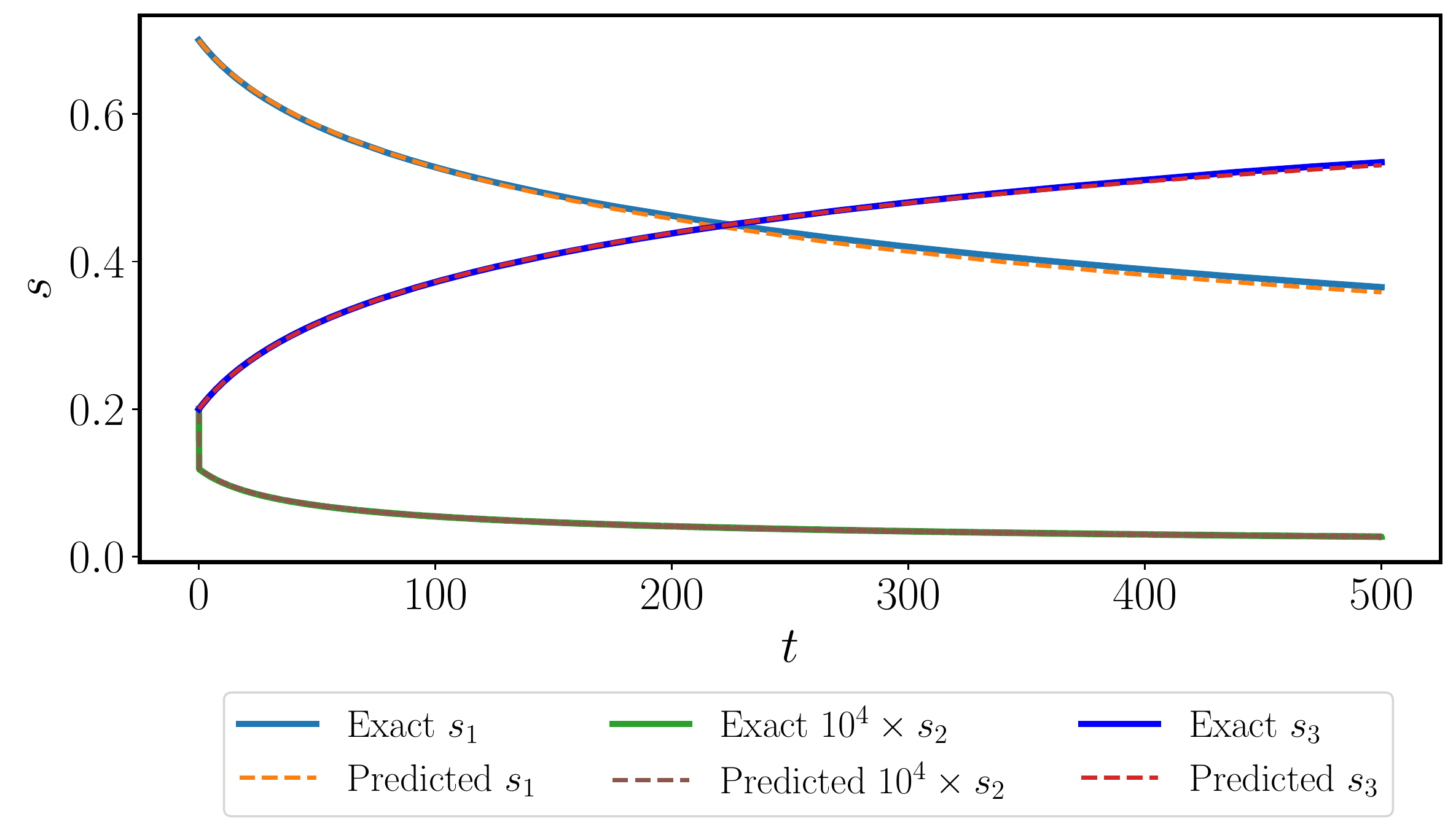}
         \label{fig: PI_deeponet_stiff_ODE_s_pred_3}
     \end{subfigure}
        \caption{{\em Stiff ODE:} Predicted solutions of a trained physics-informed DeepONet using Algorithm \ref{alg: long_time_integration} for three different initial conditions.}
        \label{fig: PI_deeponet_stiff_ODE_s_pred_examples}
\end{figure}

\clearpage
\subsection{Wave propagation}

\begin{figure}[h]
    \centering
    \includegraphics[width=0.4\textwidth]{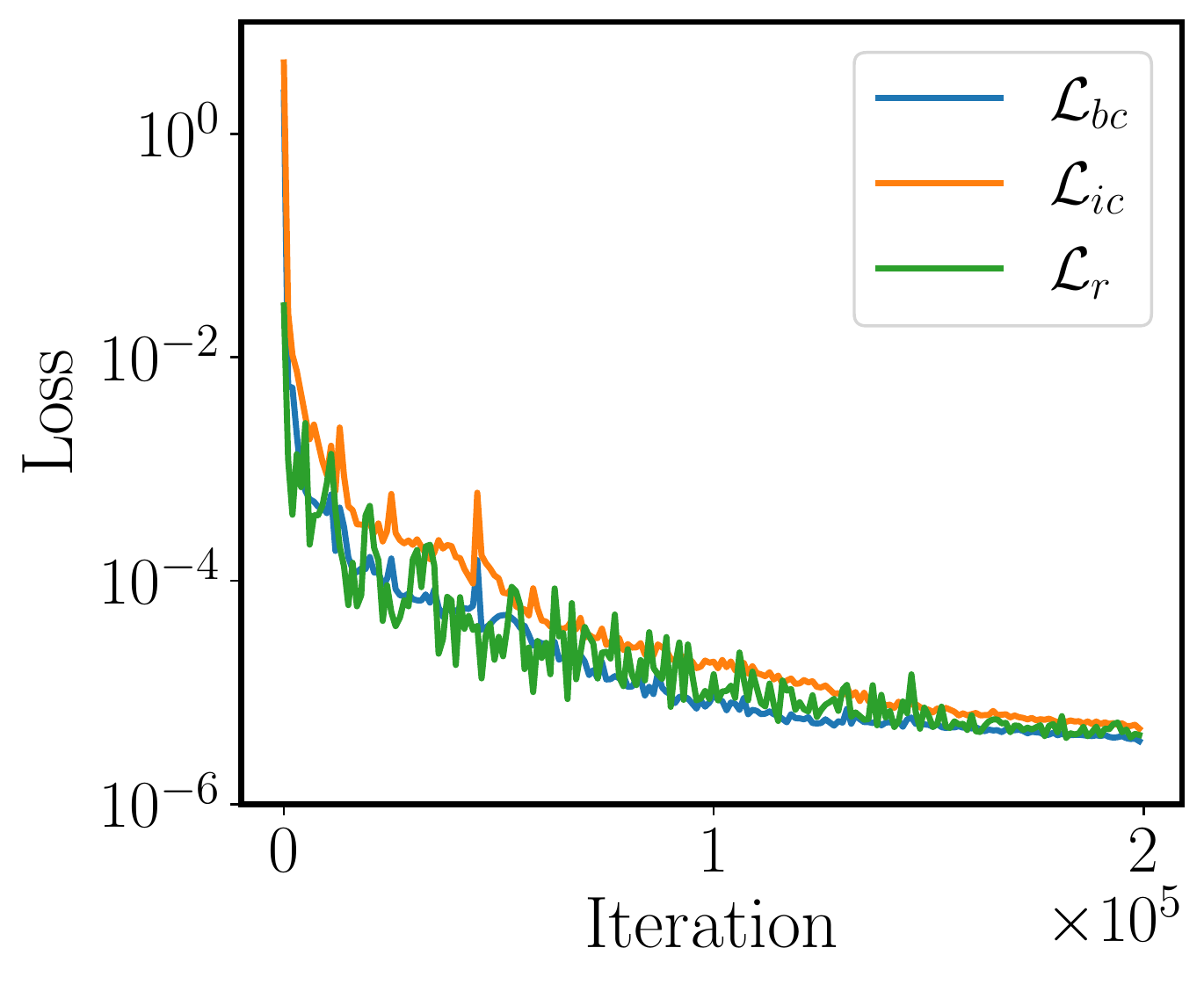}
    \caption{{\em Wave equation:}  Training loss convergence of a   physics-informed DeepONet model for $2 \times 10^5$ iterations of gradient descent using the Adam optimizer. }
    \label{fig: PI_deeponet_wave_loss}
\end{figure}

\clearpage
\subsection{Diffusion-reaction dynamics}

\begin{figure}[h]
    \centering
    \includegraphics[width=0.4\textwidth]{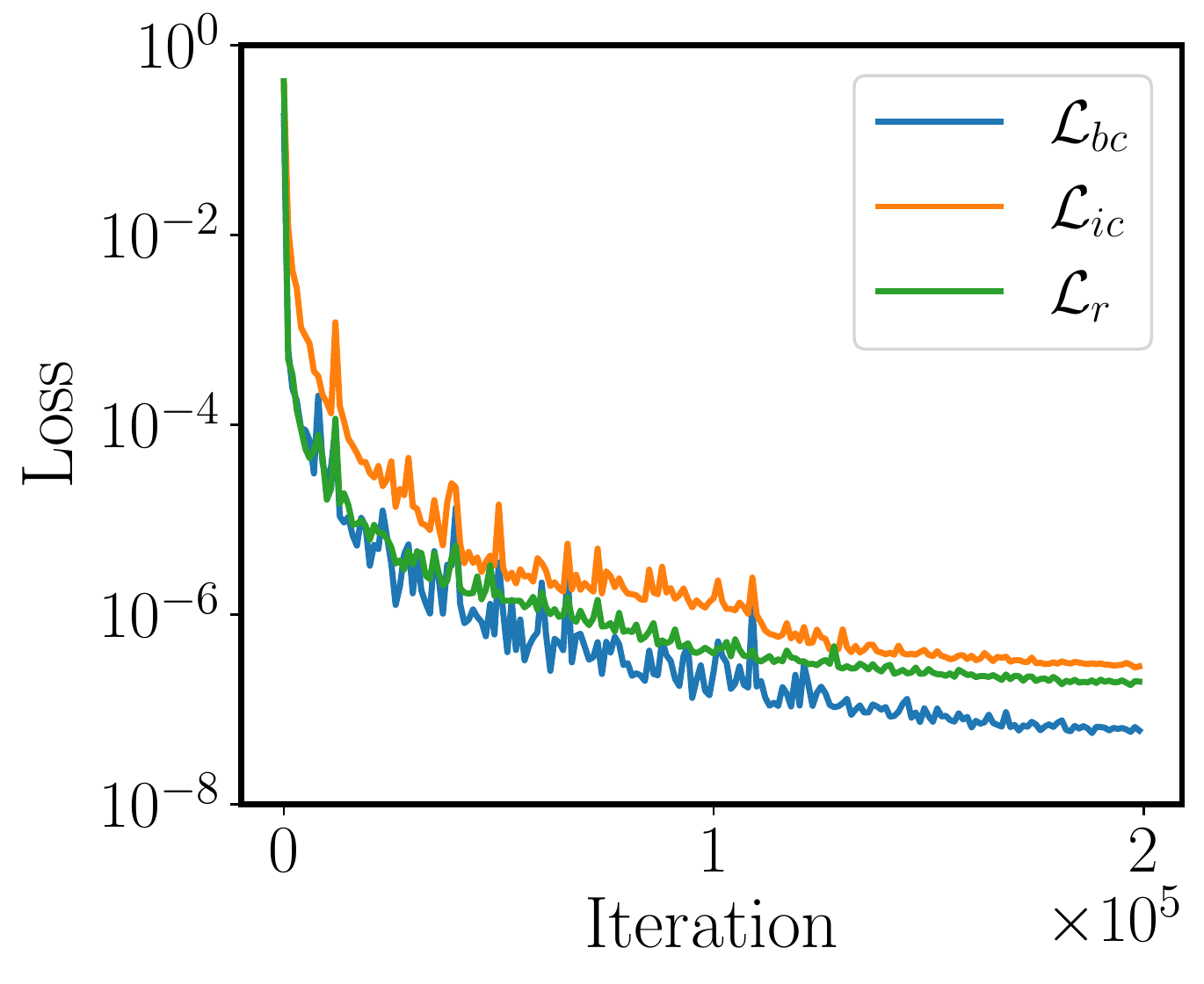}
    \caption{{\em Diffusion-reaction system:}  Training loss convergence of a physics-informed DeepONet model for $2 \times 10^5$ iterations of gradient descent using the Adam optimizer. }
    \label{fig: PI_deeponet_DR_loss}
\end{figure}

\begin{figure}[h]
     \centering
     \begin{subfigure}[b]{0.7\textwidth}
         \centering
         \includegraphics[width=\textwidth]{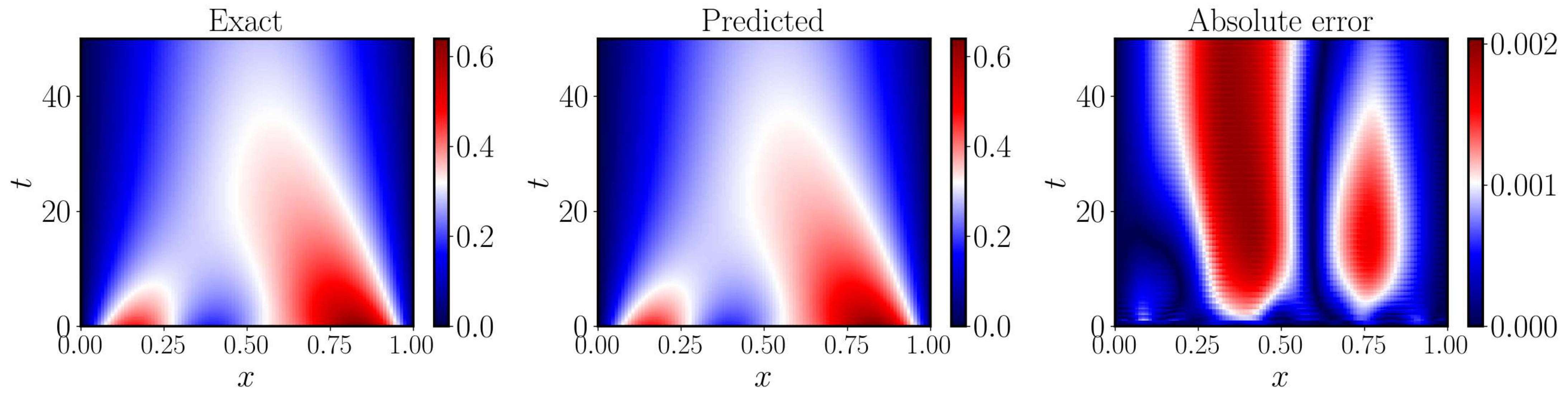}
         \label{fig: PI_deeponet_DR_s_pred_1}
     \end{subfigure}
          \begin{subfigure}[b]{0.7\textwidth}
         \centering
         \includegraphics[width=\textwidth]{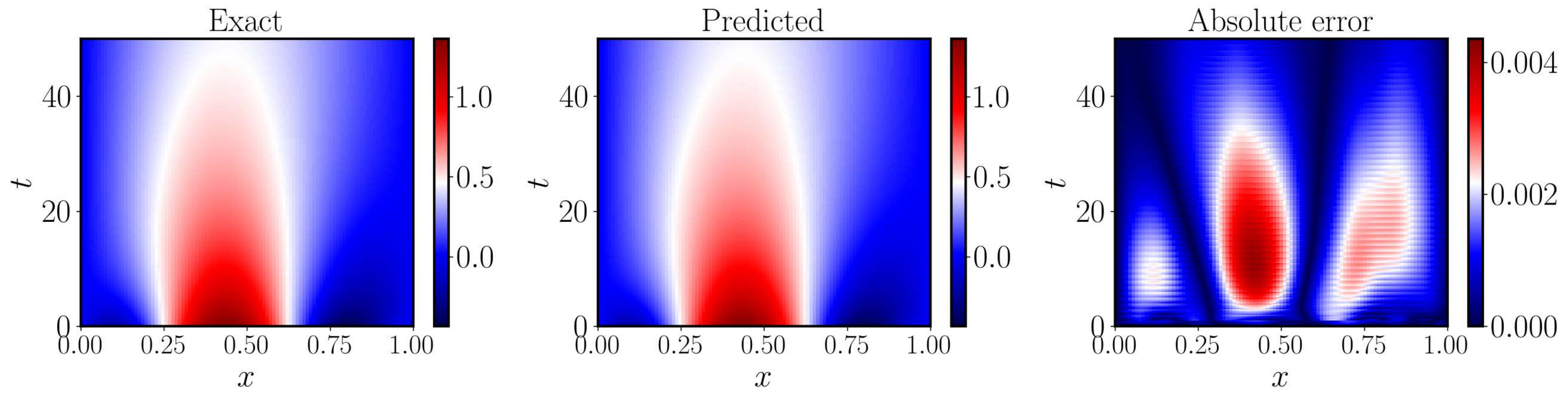}
         \label{fig: PI_deeponet_DR_s_pred_2}
     \end{subfigure}
               \begin{subfigure}[b]{0.7\textwidth}
         \centering
         \includegraphics[width=\textwidth]{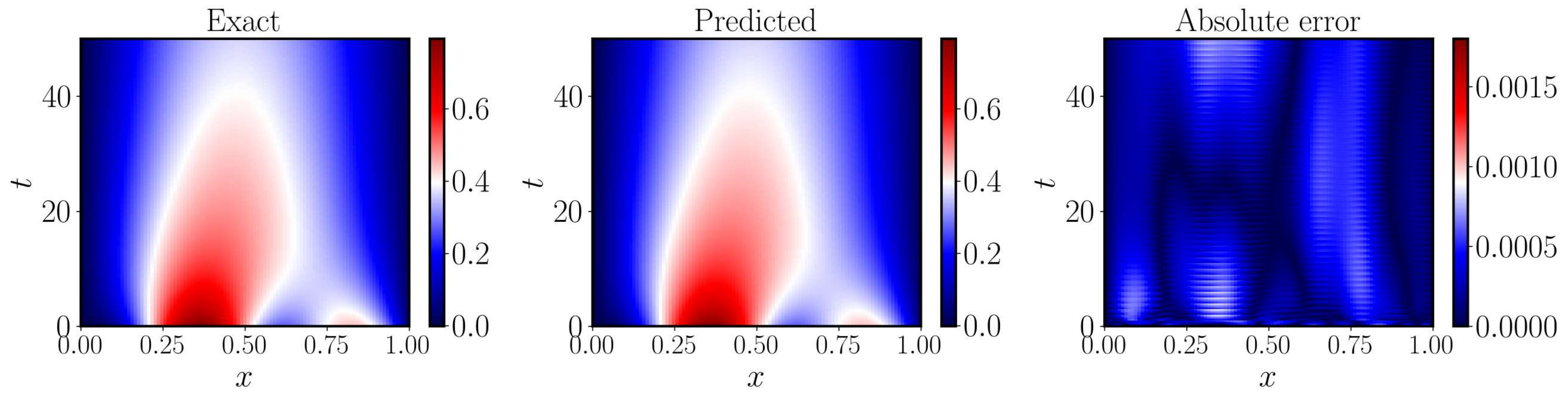}
         \label{fig: PI_deeponet_DR_s_pred_3}
     \end{subfigure}
        \caption{{\em Diffusion-reaction system:} Predicted solutions of a trained physics-informed DeepONet using Algorithm \ref{alg: long_time_integration} for three different initial conditions.}
        \label{fig: PI_deeponet_DR_s_pred_examples}
\end{figure}

\clearpage
\subsection{KDV equation}

\begin{figure}[h]
     \centering
     \begin{subfigure}[b]{0.4\textwidth}
         \centering
    \includegraphics[width=\textwidth]{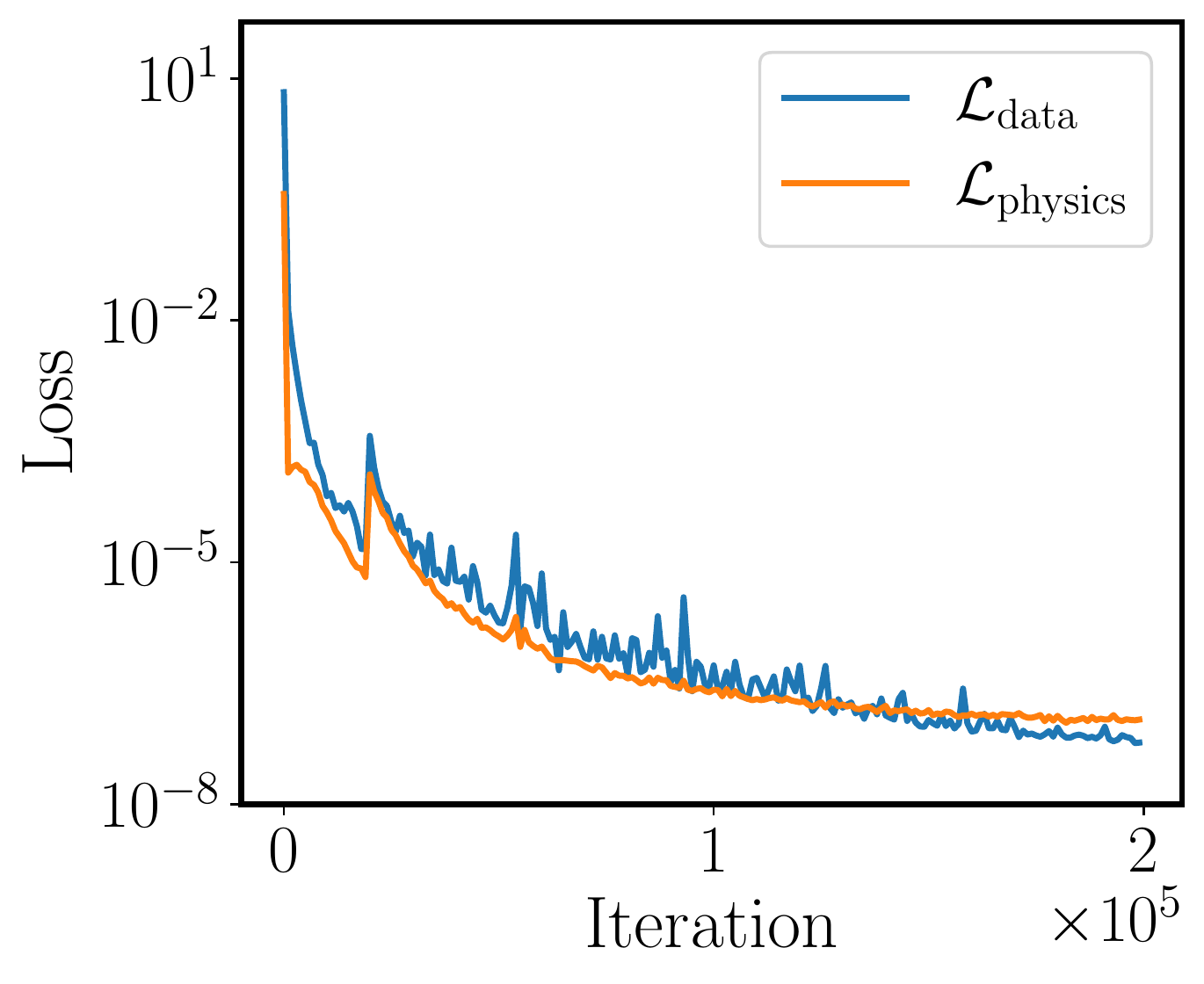}
         \label{fig: PI_deeponet_KDV_loss}
     \end{subfigure}
     \begin{subfigure}[b]{0.4\textwidth}
         \centering
    \includegraphics[width=\textwidth]{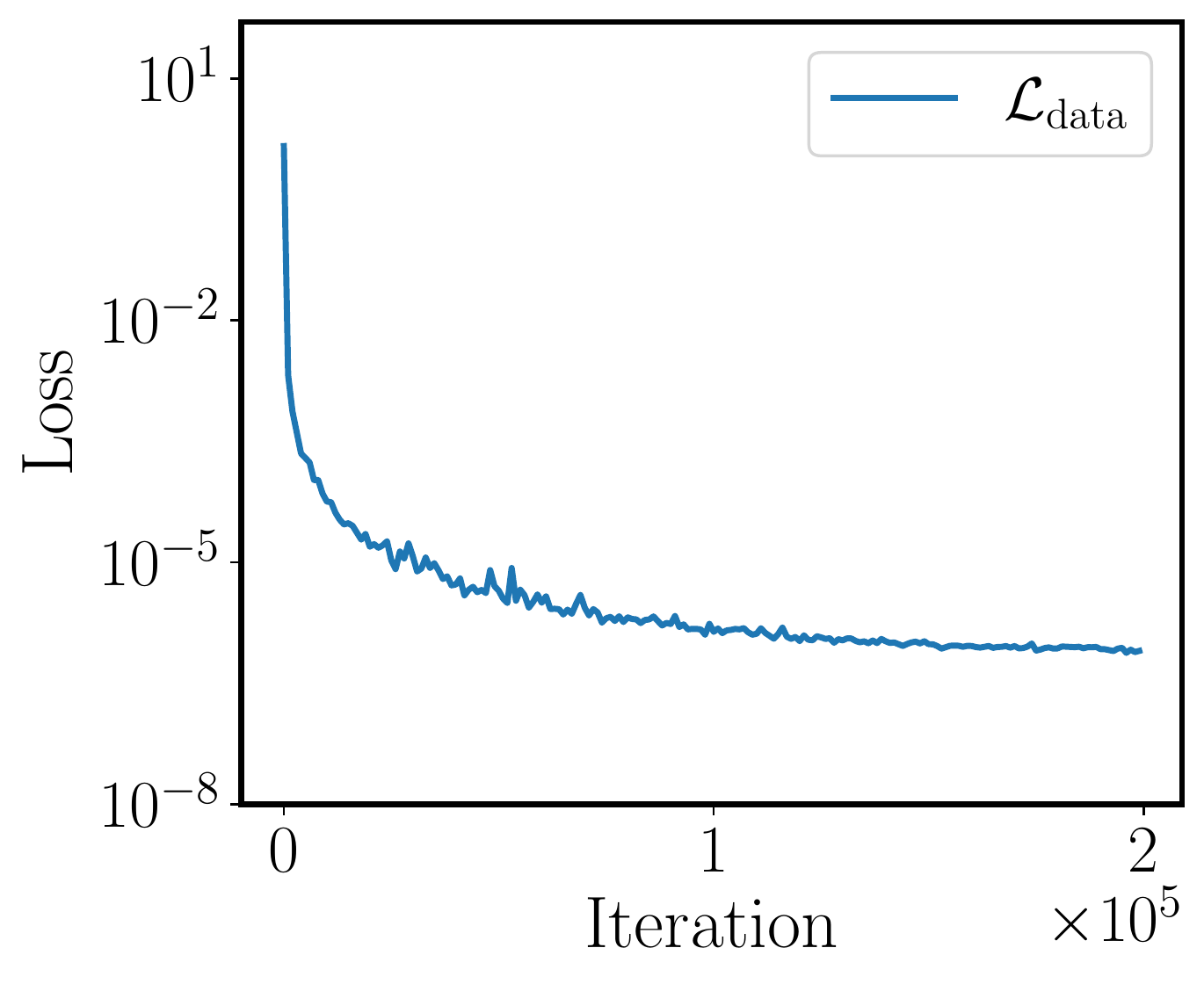}
         \label{fig: deeponet_KDV_loss}
     \end{subfigure}
       \caption{{\em Stiff ODE:} {\em Left:} Training loss convergence of a  physics-informed DeepONet model for $2 \times 10^5$ iterations of gradient descent using the Adam optimizer. {\em Right:}  Training loss convergence of a conventional DeepONet model for $2 \times 10^5$ iterations of gradient descent using the Adam optimizer.}
        \label{fig: KDV_losses}
\end{figure}